\documentclass[runningheads]{llncs}
\usepackage[english]{babel}
\usepackage{graphicx}
\usepackage{subcaption}
\usepackage{hyperref}
\usepackage{algorithm}
\usepackage{algpseudocode}
\usepackage{amsmath}
\usepackage{pifont}
\usepackage{stfloats}
\usepackage[autostyle]{csquotes}
\usepackage{tabularray}
\usepackage{comment}
\usepackage[sorting=none]{biblatex}
\newcommand{\cmark}{\ding{51}} 
\newcommand{\xmark}{\ding{55}} 
\usepackage{enumitem}
\usepackage{amssymb}
\usepackage{threeparttable}

\begin{document}
\title{From Gaming to Research: GTA V for Synthetic Data Generation for Robotics and Navigations}
\titlerunning{GTA V for Synthetic Data Generation for Robotics and Navigations}
%
%
\author{Matteo Scucchia, Matteo Ferrara, Davide Maltoni}
%
%
\institute{Department of Computer Science and Engineering, University of Bologna, Italy \email{matteo.scucchia2@unibo.it}, \email{matteo.ferrara@unibo.it}, \email{davide.maltoni@unibo.it}}

\maketitle              
\begin{abstract}
In computer vision, the development of robust algorithms capable of generalizing effectively in real-world scenarios more and more often requires large-scale datasets collected under diverse environmental conditions. However, acquiring such datasets is time-consuming, costly, and sometimes unfeasible. To address these limitations, the use of synthetic data has gained attention as a viable alternative, allowing researchers to generate vast amounts of data while simulating various environmental contexts in a controlled setting. In this study, we investigate the use of synthetic data in robotics and navigation, specifically focusing on Simultaneous Localization and Mapping (SLAM) and Visual Place Recognition (VPR). In particular, we introduce a synthetic dataset created using the virtual environment of the video game Grand Theft Auto V (GTA V), along with an algorithm designed to generate a VPR dataset, without human supervision. Through a series of experiments centered on SLAM and VPR, we demonstrate that synthetic data derived from GTA V are qualitatively comparable to real-world data. Furthermore, these synthetic data can complement or even substitute real-world data in these applications.
This study sets the stage for the creation of large-scale synthetic datasets, offering a cost-effective and scalable solution for future research and development.

\end{abstract}
\section{Introduction}
In the context of computer vision, numerous tasks require extensive datasets to train robust algorithms capable of generalizing effectively in real-world scenarios. Techniques based on Deep Learning necessitate substantial data volumes, often acquired under diverse environmental conditions (especially in outdoor scenarios) to ensure adaptability to varying contexts, including lighting conditions, weather variations, seasonal changes, and human interactions. The acquisition of such datasets poses hard challenges: it is a laborious task, time- and energy-consuming, often also costly, and not always feasible. 

To address these challenges, the generation of synthetic data has emerged as a promising solution. By creating data synthetically, researchers can simulate diverse environmental conditions in a controlled manner and generate vast amounts of data, thus substituting or integrating real-world datasets for algorithm training and improvement.
For this purpose, synthetic data must be qualitatively close to real data, and this is not trivial, since real-world data can be difficult to replicate in a simulation, being subject to complex physical rules.

The aim of this study is to demonstrate that synthetic data obtained from the famous video game Grand Theft Auto V (GTA V) \cite{gtav} can be used in conjunction with or even in place of real datasets in robotics and navigation tasks. For this purpose, data were acquired from GTA V for Simultaneous Localization and Mapping (SLAM) and Visual Place Recognition (VPR). Several experiments were conducted to demonstrate the usability of these synthetic data in both tasks. The advantages of these acquisition methods are remarkable. For instance, a notable challenge in robotics is the long-term or lifelong SLAM \cite{openloris}. This refers to a robot's ability to deal with very large and long-term explorations, where it can repeatedly traverse the same path under varying conditions, with the environment evolving over time. To achieve accurate pose estimation, the robot must relocalize itself when revisiting the same locations. This problem is known as Loop Closure Detection (LCD) or relocalization \cite{lcd_survey}. To the best of our knowledge, there are very few existing SLAM datasets in which a robot retraces the same paths multiple times over many hours, under different lighting and weather conditions, within a single acquisition. This would require extensive effort in the real world, but it can be easily achieved by leveraging GTA V. Furthermore, in a simulated environment when a path is retraced we can introduce local pose perturbation to adds variability in the camera view and make LCD more challenging.

Even if the idea of acquiring datasets from simulators and games like GTA V is not new, in this work we specifically focuses on synthetic data generation for robotic tasks, where real data are still predominantly used, particularly for VPR. Furthermore, we developed a simple but effective automatic algorithm for generating a VPR dataset from GTA V acquisitions, eliminating the need for human supervision. 


This work paves the way for the creation of unprecedentedly large datasets for all the aforementioned robotic vision tasks, composed solely of synthetic data from GTA V, which can be easily and rapidly enriched at no cost. The main novelties and contributions of this work are:

\begin{itemize}
    \item The creation of a novel synthetic dataset acquired from GTA V, specifically designed for use in robotics and navigation tasks.
    \item The development of an autonomous frame selection algorithm to generate a VPR dataset from synthetic data acquired with GTA V.
    \item The description of the pipeline and tools used to create the dataset to allow other researcher to easily generate further datasets tuned for their needs.
    \item A series of experiments to support and validate the use of synthetic data acquired from GTA V for SLAM and VPR applications.
\end{itemize}
Code and data will be released at \href{https://github.com/scumatteo/GTA-rn}{https://github.com/scumatteo/GTA-rn} upon acceptance.

The rest of the paper is organized as follows: Section \ref{background} introduces the data that can be extracted from GTA V and the computer vision tasks that can be addressed utilizing such data. Section \ref{related} discusses existing SLAM and VPR datasets and presents other datasets generated using GTA V. Section \ref{capture} details the generation pipeline, while Section \ref{preliminary} and \ref{vpr_dataset} present the dataset and the automatic frame selection algorithm for VPR. Section \ref{experiments} reports the experiments conducted for VPR and SLAM, along with key insights. Finally, Section \ref{conclusion} provides conclusions and outlines future work.

\section{Background} \label{background}
With the rapid growth of deep learning, artificial neural networks have begun to be used to solve a wide variety of computer vision tasks, with outstanding performance. To train deep neural networks, researchers started to acquire large datasets, usually task-specific.

\subsection{Type of data} \label{data}
While some of the data commonly used in computer vision tasks can be easily acquired in the real world through fairly inexpensive sensors (e.g., RGB images), the acquisition of other data (e.g., 3D point clouds) requires expensive devices. Other data, such as bounding boxes or segmented images, require manual labeling or the use of semi-automated software tools, leading to a significant investment of time and resources. Hence, the opportunity to create highly accurate and realistic data without any expense by utilizing GTA V offers a remarkable benefit.\\

\textbf{RGB image}: an RGB image uses the RGB color model to represent colors by combining red, green, and blue channels. Each pixel is represented by a triplet (R, G, B), with intensity values from 0 (no intensity) to 255 (full intensity). In GTA V it is the image captured from the screen during the game.\\

\begin{figure}[ht!]
    \centering
    
    \begin{subfigure}[b]{0.48\textwidth}
        \centering
        \includegraphics[width=\textwidth]{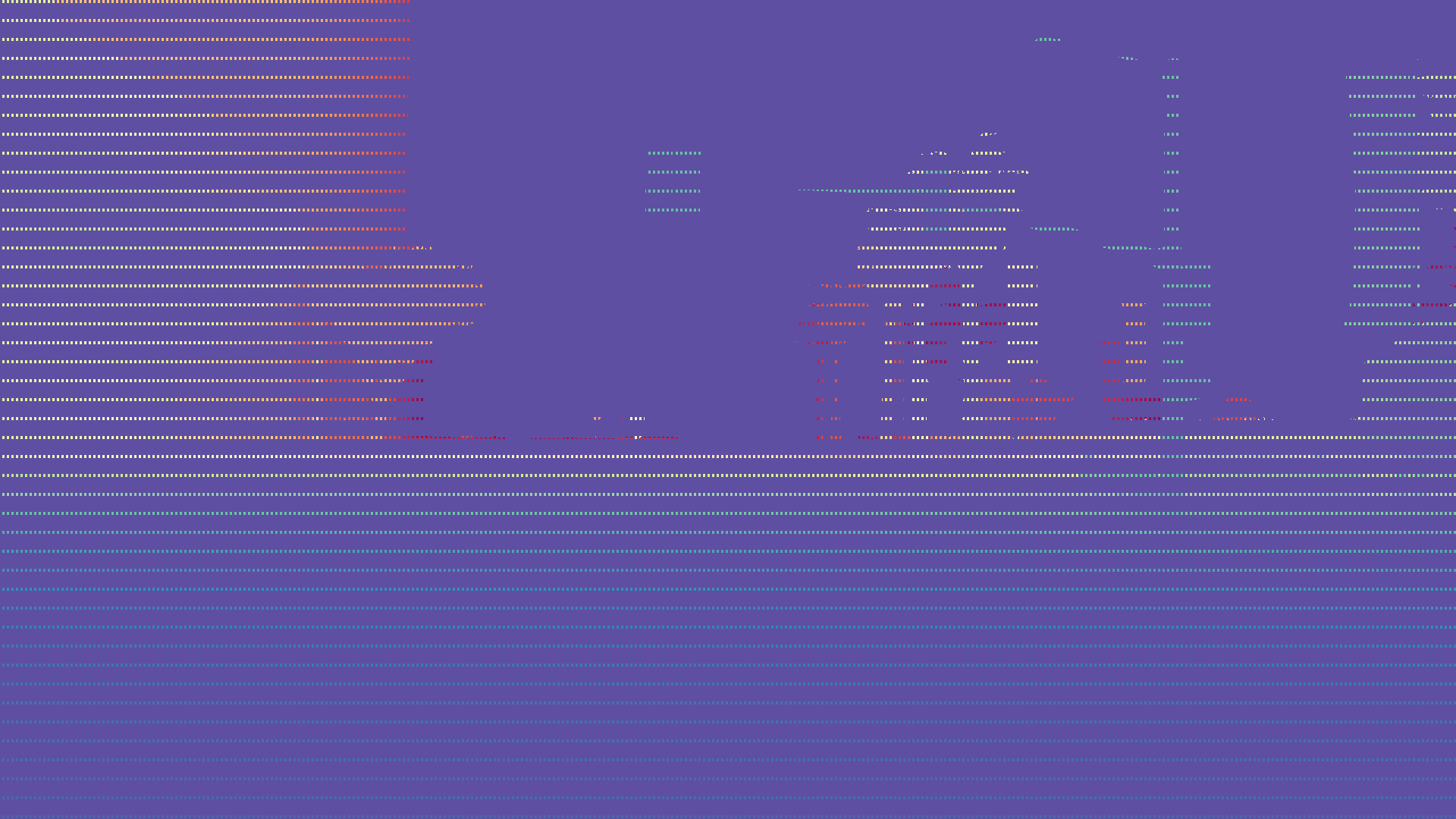}
        \caption{}
        \label{fig:real_sparse_depth}
    \end{subfigure}
    \hfill
    \begin{subfigure}[b]{0.48\textwidth}
        \centering
        \includegraphics[width=\textwidth]{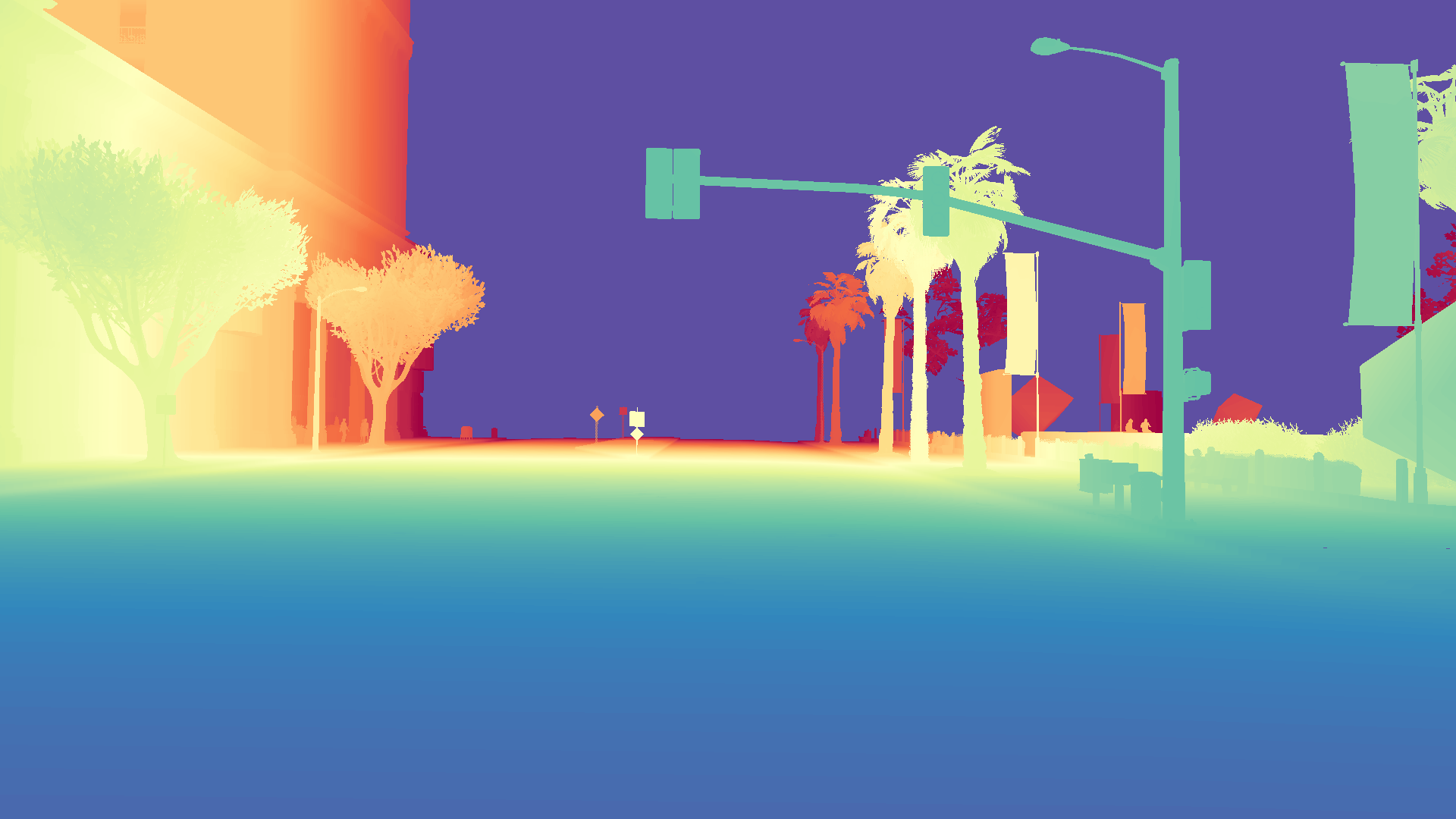}
        \caption{}
        \label{fig:real_depth}
    \end{subfigure}
    
    \caption{ On the left, (a) shows a sparse depth map, while on the right, (b) is the corresponding dense depth.}
    \label{fig:real_depth_images}
\end{figure}

\textbf{Depth image}: also known as depth map, it is a type of image that encodes the distance from the sensor to the objects in the scene. Unlike a regular RGB image that captures color information, a depth image captures spatial information, representing the three-dimensional structure of the scene.
Usually, each pixel in a depth image represents the distance in millimeters from the objects, and it is encoded as a 16-bit single channel image, so the maximum theoretical distance is $2^{16}$mm$ \simeq 65$m, but also different encodings exist. Depth images can be dense or sparse: in fact, when acquired through sensors, for not all the pixels the depth can be correctly computed, due to occlusions or non-reflective surfaces, resulting in holes in the depth map. The sensor can be (i) a camera (such as a depth camera) which captures both RGB and depth images (the so called RGB-D images) and in this case the depth map is typically dense, as the largest part of the pixels is correctly acquired (Fig. \ref{fig:real_depth}), or (ii) a laser used to make depth maps by projecting on a plane the 3D points acquired and the result is typically sparse (Fig. \ref{fig:real_sparse_depth}). An advantage of using data extracted from GTA is that the depth images are dense and complete, without holes. This is because the depths are calculated and constructed by the game during the rendering pipeline, not acquired through sensors, so they are accurate without any error or noise. GTA V uses an inverse logarithmic encoding in which values are inverted (smaller values indicate greater distances) on a logarithmic scale in the range 0-1, with a maximum distance of $960$m. 
\\

\begin{figure}[ht!]
    \centering
    \includegraphics[width=\textwidth]{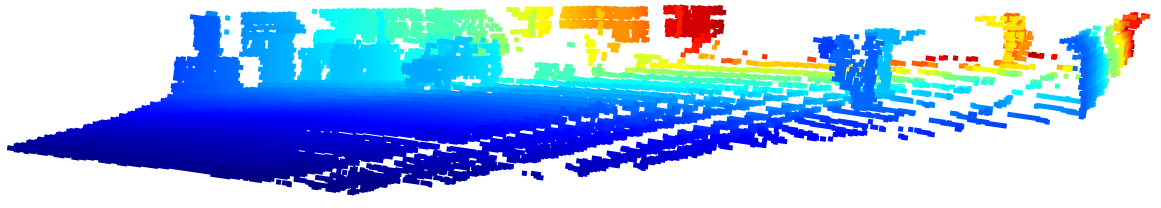}
    \caption{Three-quarter view of a point cloud acquired through a 3D laser from the KITTI dataset \cite{kitti}. Color fades from blue to red, indicating greater proximity to the sensor at blue and greater distance at red. }
    \label{fig:pcd}
\end{figure}

\textbf{Point cloud}: it can be seen as an RGB-D image but projected in a 3D scene. Each pixel becomes a point (with eventually a color) in the 3D world, at a particular distance from the sensor (Fig \ref{fig:pcd}). A point cloud is typically acquired through lasers and can be converted to a depth image and vice-versa. \\

\textbf{Camera pose}: it is a vector of six elements $[x, y, z, \phi_x, \phi_y, \phi_z]$ which represents the translation ($x, y, z$) and the rotation ($\phi_x, \phi_y, \phi_z$) of the camera with respect to the world reference frame. Camera pose is very useful in providing ground truth for tasks such as SLAM, where knowledge of precise camera pose is required. Obtaining highly accurate camera position information is a challenging and expensive task in the real world, while it is totally free and accurate using GTA V. \\

\begin{figure}[ht!]
    \centering
    \includegraphics[width=0.8\textwidth]{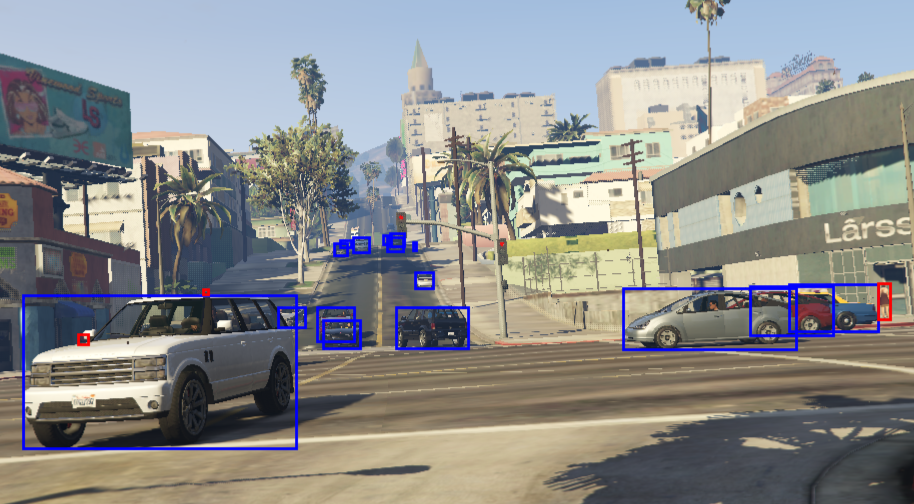}
    \caption{Image with bounding boxes from the PreSIL dataset \cite{presil}, acquired from GTA V. The blue rectangle are the bounding boxes for the vehicles, while the red ones are for the pedestrians.}
    \label{fig:obj}
\end{figure}

\textbf{Bounding boxes}: they are rectangular boxes used to define the location of objects within an image. They serve as the primary method for identifying and localizing objects in computer vision tasks, such as object detection, autonomous driving, surveillance systems and so on. Bounding boxes are typically represented using the coordinates of top-left $(x_1, y_1)$ and bottom-right $(x_2, y_2)$ corners, or using the coordinates of the center point $(c_x, c_y)$ with the width $w$ and the height $h$ of the rectangle. Fig. \ref{fig:obj} shows an example of image with bounding boxes. \\

\begin{figure}[ht!]
    \centering
    
    \begin{subfigure}[b]{0.48\textwidth}
        \centering
        \includegraphics[width=\textwidth]{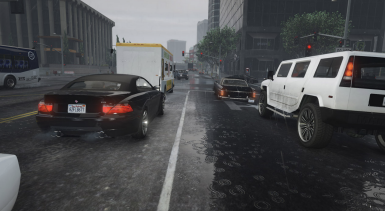}
        \caption{}
        \label{fig:rgb}
    \end{subfigure}
    \hfill
    \begin{subfigure}[b]{0.48\textwidth}
        \centering
        \includegraphics[width=\textwidth]{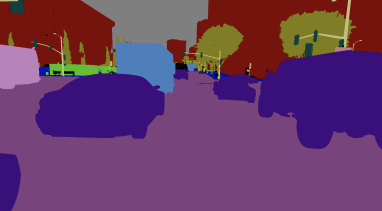}
        \caption{}
        \label{fig:segmented}
    \end{subfigure}
    \caption{ On the left, (a) shows an RGB image, while on the right (b) shows the corresponding segmented image. The images are from the GTA5 dataset \cite{gta_seg}, acquired from GTA V. Each pixel is colored according to the category of the object to which it belongs.}
    \label{fig:segmentation}
\end{figure}

\textbf{Segmented images}: in the context of computer vision and image processing, they are images that have been divided into different segments or regions, each corresponding to different objects, or other meaningful areas within the image. Segmented images are commonly used in (i) autonomous driving, to understand the surroundings by segmenting roads, pedestrians, vehicles, and other objects, (ii) robotics, to help robots understand and interact with the environment by identifying objects and their locations, (iii) medical imaging, to identify different anatomical structures in medical scans (e.g., tumors) and so on. Fig \ref{fig:segmentation} shows an example of a segmented image.\\

Additional data can also be captured by GTA V, such as movement speed, time of day, and stencil images. However, these are not important for the scope of this paper.

\subsection{Tasks} \label{tasks}
Here we report the main tasks of the computer vision community that can be addressed using data acquired from GTA V. \\

\textbf{Place recognition}: it involves identifying previously visited places by comparing sensor data (e.g., images, laser scans) to a database of known locations. It is a difficult task because places can change their appearance over time due to changing lighting conditions (e.g., day/night), seasonality, occlusions (people, trees, cars, etc.), or changes in viewpoint.

If place recognition relies on images, the task takes the name of Visual Place Recognition (VPR) and can be seen as an image retrieval problem, in which the image of a place is compared with images stored in a database to retrieve the most similar places. To measure the similarity between two images, several methods have been proposed in the literature \cite{image_sim}. It generally requires the extraction of numerical feature vectors from the images through hand-crafted methods or trained deep neural networks. Once the feature vectors have been computed, it is possible to apply a similarity/distance metric (e.g., cosine similarity) to find the most similar images in the database.   \\

\textbf{SLAM}: Simultaneous Localization and Mapping (SLAM) is a robotic problem in which a mobile robot navigates in an unfamiliar environment and, perceiving it through a series of sensors, the robot must create a map, locating itself on the map under construction \cite{slam_part1}. This is a critical task to give robots autonomous navigation capabilities, especially when GPS cannot provide good performance, such as in indoor scenarios. There are several approaches in the literature for this problem, the current state-of-the-art solutions fall into the graph-based category, in which each robot pose is a node in a graph representing the map, and measurements of displacements between different sensor acquisitions are edges, which represent constraints \cite{graph_slam}. This graph can be optimized when the robot returns to places it has already visited and recognizes them as known, so SLAM is closely linked with place recognition. This process takes the name of (i) Loop Closure Detection (LCD) when the robot ``closes a loop'' during the navigation or (ii) Relocalization when the robot returns to previously visited places in successive explorations \cite{lcd_survey}.
The most common solutions typically rely on lasers, cameras and inertial sensors (accelerometer and gyroscope) and depending on the sensors used, different types of maps can be built by the SLAM algorithm. In this work, we use a visual SLAM solution, in which images are used to compute visual odometry \cite{visual_odometry}, i.e., pose estimation based on the robot's motion.\\

Other computer vision tasks can be addressed using data from GTA V:
\begin{itemize}
    \item \textbf{Depth estimation} involves predicting depth information from RGB images to mitigate the high cost of depth sensors. However, challenges such as accurate scene structure representation and external factors like lighting persist. Supervised generative AI methods often perform better but require expensive, high-quality datasets.
    \item \textbf{Depth completion} focuses on transforming sparse or incomplete depth maps into dense ones, serving as a pre-processing step for tasks like SLAM. However, highly sparse data remains problematic.
    \item \textbf{Object detection} identifies and localizes objects in images or videos by combining bounding box prediction (localization) with class assignment (classification).
    \item \textbf{Semantic segmentation} goes further by assigning a class label to every pixel, providing a detailed, pixel-level understanding of an image compared to the bounding box approach of object detection.
\end{itemize}

\section{Related Works} \label{related}
This section presents an overview of the most common publicly available datasets for SLAM and VPR tasks.

\subsection{Place Recognition}
Table \ref{tab:place_recognition_datasets} reports well-known place recognition datasets. These datasets exhibit multiple illuminations and/or seasonal conditions. Note that none of them are synthetic.

\begin{table}[ht!]
\centering
\begin{tabular}{ccccc} 
\hline
\textbf{Dataset}          & \textbf{\# images} & \begin{tabular}[c]{@{}c@{}}\textbf{Data}\\\textbf{type}\end{tabular} & \begin{tabular}[c]{@{}c@{}}\textbf{Multiple}\\\textbf{ conditions}\end{tabular} & \textbf{Synthetic}   \\
\hline
\textbf{Tokyo 24/7} \cite{tokyo}                             &  76000         &  Mono RGB images           &  \cmark              &  \xmark                                                                      \\ \hline
\textbf{Alderley} \cite{alderley}                             &  29214         &  Mono RGB images           &  \cmark              &  \xmark                                                                      \\ \hline
\textbf{Nordland Dataset} \cite{nordland}                    &      28865             &    Mono RGB images         &    \cmark            &   \xmark                                                                     \\ \hline
\textbf{Pittsburgh 250k} \cite{pittsburgh}        &               $\sim$256000             &     Mono RGB images         &       \cmark         &   \xmark                                                                        \\ \hline
\textbf{Mapillary} \cite{mapillary}               &      $\sim$1.6 million            &     Mono RGB images        &      \cmark          &     \xmark                                                                   \\ \hline
\end{tabular}
\caption{The main publicly available place recognition datasets.}
\label{tab:place_recognition_datasets}
\end{table}

\subsection{SLAM} \label{slam_datasets}
Table \ref{tab:slam_datasets} lists well-known and widely used SLAM datasets. All the datasets reported are provided with ground truth. In recent years, several synthetic datasets have been proposed for SLAM, in particular, GTASynth \cite{gtasynth} is another dataset extracted from GTA V. In the table below, IMU stands for Inertial Measurement Unit and refers to inertial sensors such as accelerometers and gyroscopes.

\begin{table}[ht!]
\centering
\begin{threeparttable}
\begin{tabular}{ccc} 
\hline
\textbf{Dataset}               & \begin{tabular}[c]{@{}c@{}}\textbf{Data}\\\textbf{type}\end{tabular} & \textbf{Synthetic}  \\ 
\hline
\textbf{KITTI odometry} \cite{kitti}      & \begin{tabular}[c]{@{}c@{}}Stereo RGB images\\Point clouds\end{tabular}                              &         \xmark            \\ 
\hline
\textbf{TUM RGB-D} \cite{tum}            & RGB-D images                                                                             & \xmark           \\ 
\hline
\textbf{EuRoC MAV}  \cite{euroc}           & \begin{tabular}[c]{@{}c@{}}Stereo RGB images\\IMU\end{tabular}                                & \xmark         \\ 
\hline
\textbf{Newer College Dataset} \cite{newer} & \begin{tabular}[c]{@{}c@{}}Stereo RGB images\\Point clouds\\IMU\end{tabular}                   & \xmark          \\ 
\hline
\textbf{NCLT Dataset} \cite{nctl}         & \begin{tabular}[c]{@{}c@{}}360 RGB images\\Point clouds\\IMU\\Odometry\end{tabular}                     &       \xmark              \\ 
\hline
\textbf{St. Lucia Dataset} \cite{stlucia}         & \begin{tabular}[c]{@{}c@{}}Mono RGB images\\GPS\end{tabular}                     &       \xmark              \\ 
\hline
\textbf{USyd Campus Dataset}\tnote{1} \cite{usyd}         & \begin{tabular}[c]{@{}c@{}}Mono RGB images\\Point clouds\\GPS\\Odometry\end{tabular}                     &       \xmark              \\ 
\hline
\textbf{Virtual KITTI 2} \cite{virtualkitti2}      & RGB-D images                                                                                       &               \cmark      \\ 
\hline
\textbf{ICL-NUIM} \cite{iclnuim}             & RGB-D images                                                                                       & \cmark                   \\ 
\hline
\textbf{Redwood SLAM} \cite{redwood}         & \begin{tabular}[c]{@{}c@{}}RGB-D images\\Point clouds\end{tabular}                                & \cmark                   \\ 
\hline
\textbf{GTASynth}  \cite{gtasynth}            & Point clouds                                                                                   & \cmark                   \\
\hline

\end{tabular}
\begin{tablenotes}
\item[1] USyd does not have a ground truth.
\end{tablenotes}
\end{threeparttable}
\caption{The main publicly available SLAM datasets.}
\label{tab:slam_datasets}
\end{table}

\subsection{Other GTA V datasets}
In the community, the idea of extracting data from GTA V is not new. GTASynth \cite{gtasynth} is a SLAM dataset composed of sequences of point clouds with respective ground truth poses, HRSD \cite{hrsd} is an RGB-D dataset for monocular depth estimation, PreSIL \cite{presil} is an object detection dataset and GTA5 \cite{gta_seg} is a semantic segmentation dataset. Additionally, other datasets acquired from GTA V exist \cite{gta_drone, gta_people}. All these works have shown that GTA V data can be used in their respective tasks with results comparable to the use of real data. However, none of the above dataset is well suited for our VPR and SLAM tasks where multiple retracing of the same path is mandatory.
We focused on using GTA V RGB-D data to address robotics and navigation challenges, which are mainly solved using real data, particularly VPR. Additionally, to use the data for VPR, we post-processed them to create a specific dataset for this task, as detailed in Section \ref{vpr_dataset}.
An extension of this work is the acquisition of all available data from GTA V and then creating specific datasets for a wide range of computer vision tasks. In fact, while PreSIL is composed of several different data types (depth maps, point clouds, bounding boxes, segmented images), the others are very task-specific: GTASynth is composed solely of point clouds, and cannot be used for monocular depth estimation, while HRSD is composed of single RGB-D images that are not in a temporal sequence and therefore cannot be used for SLAM. Similarly, GTA5 is composed only of segmented images.

\section{How to capture data from GTA V} \label{capture}
\begin{figure}[ht!]
    \centering
    \includegraphics[width=\textwidth]{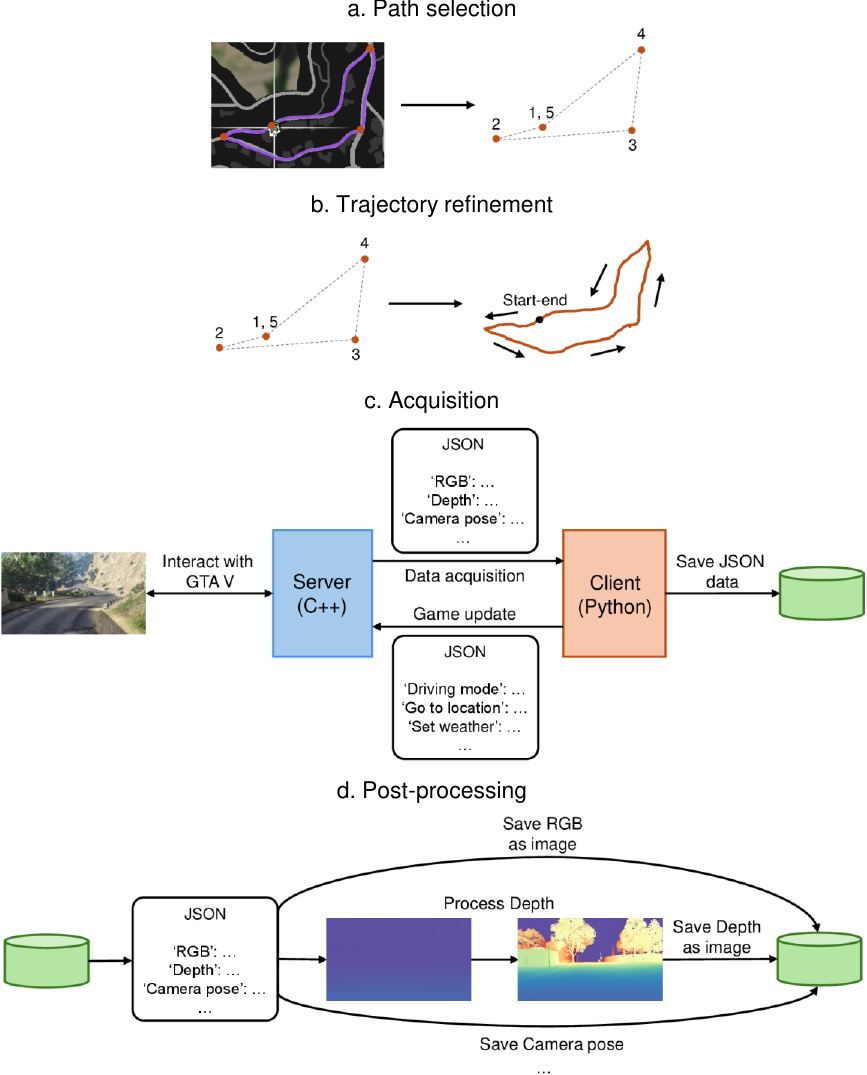}
    \caption{Data acquisition process.}
    \label{fig:process}
\end{figure}

To acquire data from GTA V we used three open-source tools: ScriptHook V \cite{scripthook}, G2D \cite{g2d} and DeepGTAV \cite{deepgtav}. ScriptHook V is a popular modding tool used primarily to modify and enhance GTA V by adding new features, custom behaviors, or entirely new content that was not originally included in the game. It allows players to run custom scripts written in the .NET framework and it works as the basis for G2D and DeepGTAV.
The idea is to use the playable character of GTA V to drive a car along a predefined path (defined using G2D) while data are recorded at a designated frame rate (using DeepGTAV).
The process can be summarized in four steps (as depicted in Fig. \ref{fig:process}):
\begin{enumerate}[label=\alph*.]
    \item \textit{Path selection}: thanks to G2D, it is possible to manually place markers on the GTA V map to outline a sparse trajectory. The order of the markers corresponds to the order in which those locations will be visited (highlighted by the vertices numbering in Fig. \ref{fig:process}a).
    \item \textit{Trajectory refinement}: a G2D script converts the sparse set of trajectory points into a dense set that follows the path of the road (see Fig. \ref{fig:process}b).
    \item \textit{Acquisition}: DeepGTAV runs a client-server architecture, used to export data from GTA V. When the acquisition starts, the game enters into a self-driving mode, and the server (written in C++) interacts with the game rendering pipeline to extract frame data and send them to the client (written in Python) as a JSON file. The client can also send messages to the server, to change the game settings (e.g., the weather, the location to reach, etc.). The client directly saves the JSON files received from the server, because processing the data is time- and computational-demanding and may prevent the acquisition at high frame rates (see Fig. \ref{fig:process}c).
    \item \textit{Post-processing}: some data (e.g., depth images) require post-processing to be stored using a standardized encoding. For this reason, once the acquisition is complete, the different types of data (i.e., RGB, depth image, point cloud, camera pose, bounding boxes and labels, segmented image) are extracted from the corresponding JSON file and stored separately for each frame (see Fig. \ref{fig:process}d).
\end{enumerate}

The tools used allow the control of various game parameters (driving mode, weather, time of day, car speed), which can be set and modified by the external client during the acquisition process. As a result, it is possible to create a simulated, fully controllable and reproducible environment with which to capture data under different conditions. This is an incredible advantage for real-world open problems, such as lifelong SLAM \cite{openloris}, for which extensive explorations, even hours or days, under different conditions are required before returning to places already visited, and such datasets do not exist in the community. Furthermore, during trajectory refinement (step b), it is possible to perturb the path and introduce small changes in the poses. This adds variability in the camera view when retracing the same path, which is a desirable condition for SLAM and VPR datasets.

At the end of the acquisition process, captured data can be selected to create various datasets useful for different tasks. For instance:
\begin{itemize}
\item RGB, depth images and poses ground truth can be used for monocular depth estimation \cite{monodepth} or SLAM.
\item RGB images captured in different weather or day/night conditions can be useful for VPR.
\item RGB images, point clouds and poses ground truth can be used for depth completion or SLAM.
\item RGB images with precise objects bounding boxes and labels, along with depth images and segmented ground truth images can be useful for object detection and image segmentation, respectively.
\end{itemize}
The capability of this acquisition approach is remarkable as it allows the collection of a wide variety of data under a large range of controllable conditions, for many different tasks, very efficiently at almost no cost.
Since the game enters into autonomous driving mode, during the acquisition process, at the moment only outdoor data can be acquired. 

\section{Acquired dataset} \label{preliminary}
\begin{figure}[ht!]
    \centering
    
    \begin{subfigure}[b]{0.35\textwidth}
        \centering
        \includegraphics[width=\textwidth]{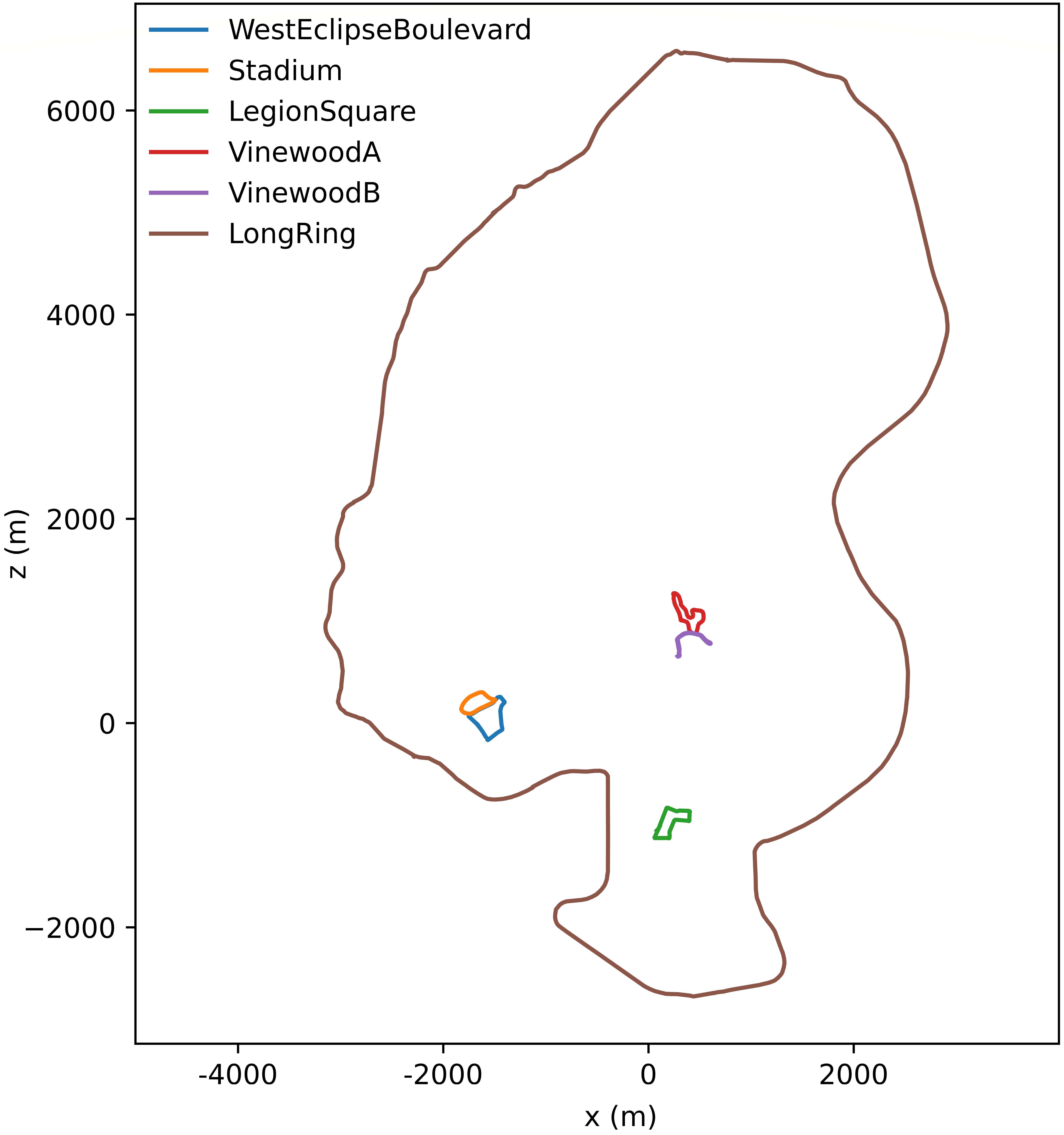}
        \vspace{0.01cm}
        \caption{}
        \label{fig:gta_paths}
    \end{subfigure}
    \hfill
    \begin{subfigure}[b]{0.3\textwidth}
        \centering
        \includegraphics[width=\textwidth]{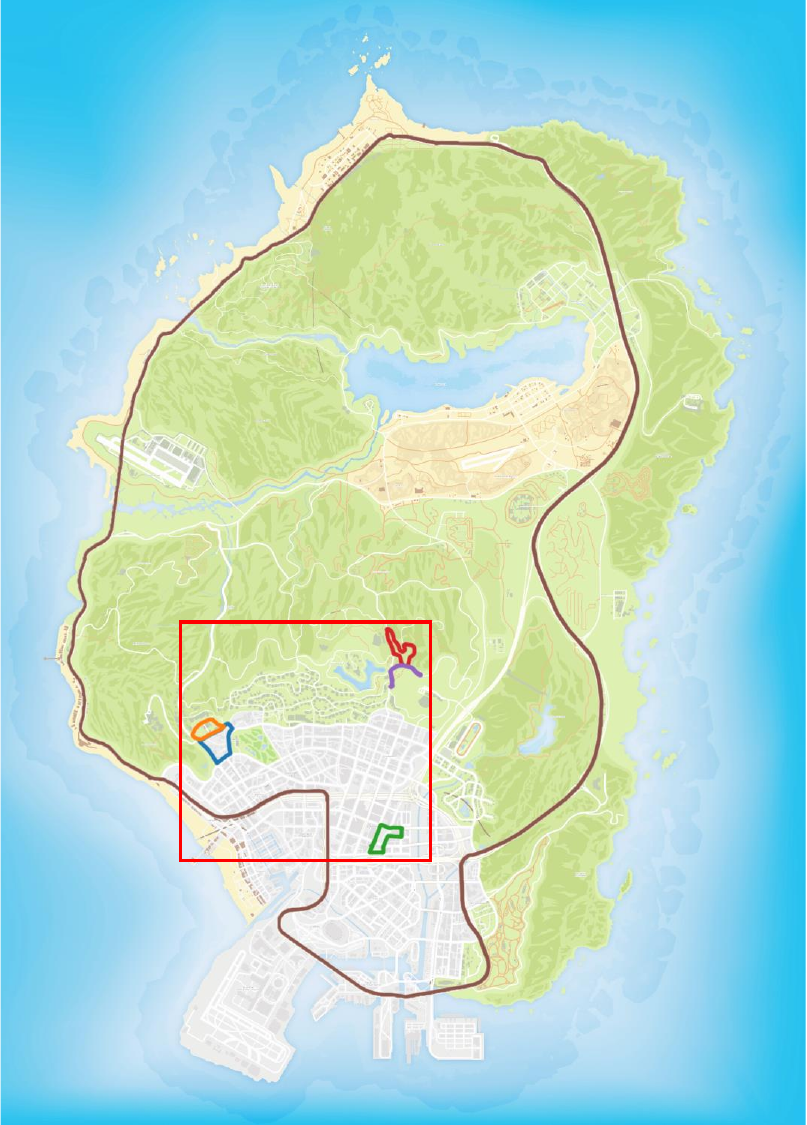}
        \caption{}
        \label{fig:gta_map_paths}
    \end{subfigure}
    \hfill
    \begin{subfigure}[b]{0.25\textwidth}
        \centering
        \includegraphics[width=\textwidth]{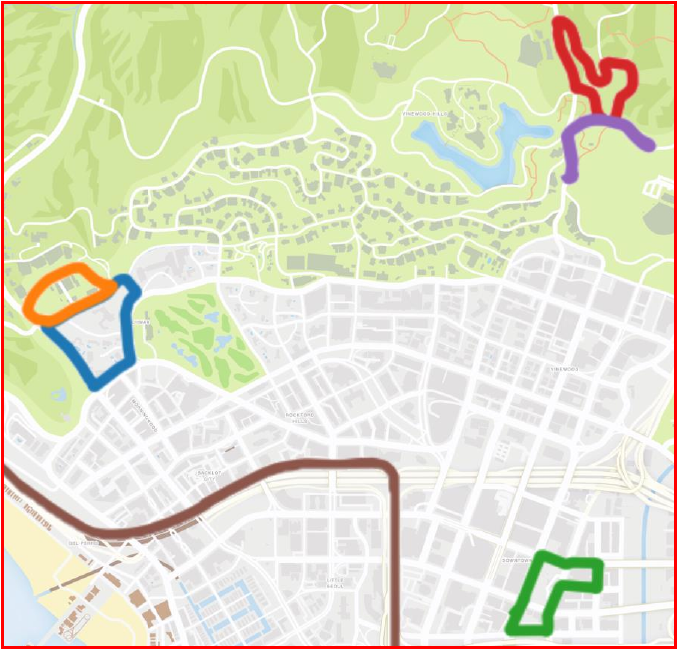}
        \vspace{0.5cm}
        \caption{}
        \label{fig:gta_map_paths_zoom}
        
    \end{subfigure}
    \caption{The six paths acquired from GTA V. On the left, (a) highlights the scale and the legend to provide a clear visualization of the paths, while on the center, (b) reports the same paths on the GTA map. The red square identifies the portion of the map zoomed in (c) to show better the smaller paths.}
    \label{fig:gta_maps}
\end{figure}

\begin{figure}[ht!]
    \centering
    
    \begin{subfigure}[b]{0.48\textwidth}
        \centering
        \includegraphics[width=\textwidth]{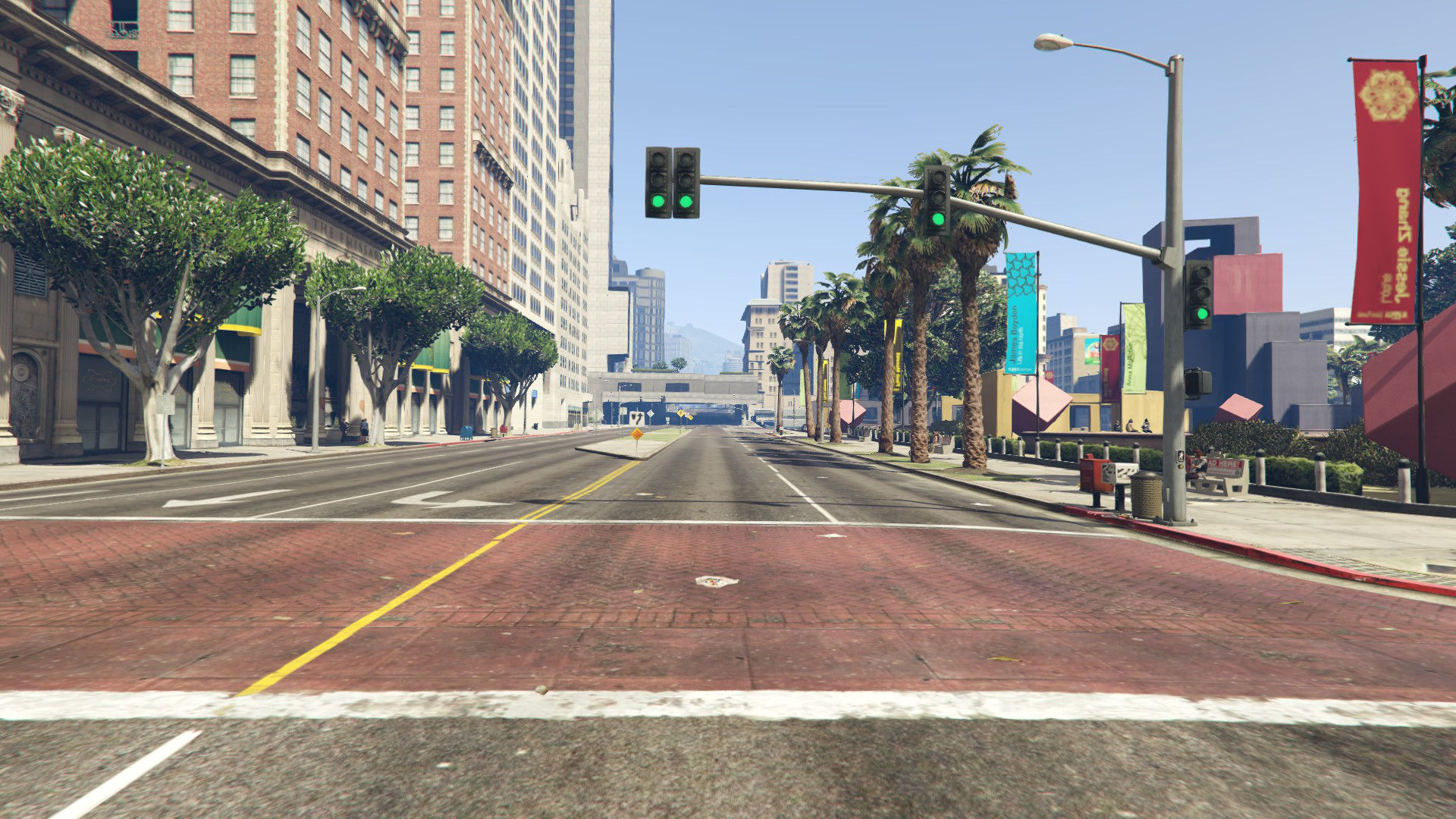}
        \caption{}
        \label{fig:gta_rgb}
    \end{subfigure}
    \hfill
    \begin{subfigure}[b]{0.48\textwidth}
        \centering
        \includegraphics[width=\textwidth]{images/gta_depth_color.png}
        \caption{}
        \label{fig:gta_depth}
    \end{subfigure}
    \caption{RGB-D image extracted from the Legion Square sequence. On the left, (a) the RGB image, on the right, (b) the corresponding post-processed depth. It is worth noting the high quality of the depth image. }
    \label{fig:gta_rgbd}
\end{figure}

Following the process described in the previous section, a dataset consisting of six different paths of varying lengths has been acquired. For each path, multiple sequences have been recorded, retracing the same path several times under different lighting and weather conditions. The lighting conditions are day and night, while the weather conditions are sunny, overcast and rainy, obtaining five combinations: day/sunny, day/overcast, day/rainy, night/clear and night/rainy\footnote{Night/overcast is ignored because there is no significant difference compared to night/clear.}.
In this first version, only RGB-D images and the corresponding ground truth poses were captured from GTA V.
The dataset was acquired at a frame rate of 10 frames per second (but also a higher/lower frame rate can be set) and it contains a total of 111,850 high-resolution RGB and depth images, each with dimensions of 1920x1080 pixels. Table \ref{tab:dataset_stats} reports the number of frames for the different sequences contained in the dataset.
Each path in the dataset provides a unique perspective of the game's broad environment, encapsulating a variety of settings, such as urban, rural and hilly landscapes (see Fig. \ref{fig:gta_maps}). While five of the six routes are relatively short, one is a very long loop that encircles the entire island of Los Santos (game island). 
Regarding the SLAM task, this dataset offers a significant advantage: it enables the capture of extensive loops under various controlled conditions. This is crucial for testing SLAM and loop closure detection algorithms, which must be robust to handle lifelong and long-term navigation scenarios. To the best of our knowledge, no existing dataset in the community includes such extensive navigation, featuring a loop closure after a long duration and under different weather and illumination conditions. Larger loops, acquired under different conditions, will be included in the future to provide the SLAM community with a dataset for training and testing truly long-term algorithms. Fig. \ref{fig:gta_rgbd} shows an RGB-D image from the Legion Square sequence. The depth image is post-processed, to obtain the classic 16-bit depth encoding.

\begin{table}[ht!]
\centering
\begin{tblr}{
  cells = {c},
  cell{1}{2} = {c=3}{},
  cell{1}{5} = {c=2}{},
  cell{1}{7} = {r=2}{},
  vline{2-3,5,7,8} = {1}{},
  vline{1-7} = {2,9}{},
  vline{8} = {2}{},
  vline{-} = {3-8}{},
  hline{1} = {2-7}{},
  hline{2,10} = {1-6}{},
  hline{3-9} = {-}{},
}
                                             & \textbf{Day }       &                   &                & \textbf{Night } &                & \textbf{Total} \\
\textbf{Path}                                & \textbf{Extrasunny} & \textbf{Overcast} & \textbf{Rain}  & \textbf{Clear}  & \textbf{Rain}  &                \\
\textbf{Long Ring}                           & 16782               & 16627             & 15257          & 20291           & 15819          & \textbf{84776} \\
\textbf{Vinewood A}                          & 1349                & 1366              & 1000           & 1196            & 985            & \textbf{5896}  \\
\textbf{Vinewood B}                          & 1209                & 1113              & 966            & 1031            & 1114           & \textbf{5433}  \\
{\textbf{West Eclispe }\\\textbf{Boulevard}} & 1387                & 1334              & 1076           & 1112            & 1077           & \textbf{5986}  \\
\textbf{Legion Square}                       & 1331                & 1425              & 1354           & 1268            & 1142           & \textbf{6520}  \\
\textbf{Stadium}                             & 695                 & 662               & 550            & 714             & 618            & \textbf{3239}  \\
\textbf{Total}                               & \textbf{22753}      & \textbf{22527}    & \textbf{20203} & \textbf{25612}  & \textbf{20755} &                
\end{tblr}
\vspace{0.1cm}
\caption{Number of frames across the different sequences of our dataset.}
\label{tab:dataset_stats}
\end{table}

\section{VPR dataset} \label{vpr_dataset}
A dataset designed for VPR should contain images of the same location taken from various viewpoints and during varying weather conditions. For this reason, we developed an \textit{automatic frame selection} algorithm to create a dataset for VPR starting from the sequences of RGB images and the corresponding pose ground truth. For clarity, the creation of the VPR dataset was divided into two distinct steps: (i) \textit{place selection}, to select only univocal places from the sequences (see Algorithm \ref{alg:place_selection}) and (ii) \textit{frame selection}, to select multiple images from the sequences that capture each place from different viewpoints or under varying weather conditions (see Algorithm \ref{alg:frame_selection}).

Each sequence acquired using the procedure described in Section \ref{preliminary} contains a variable number of triplets. Each triplet $(i,p,c)$ is composed of an RGB image ($i$), the corresponding pose ground truth ($p$), defined as a vector $[x, y, z, \phi_x, \phi_y, \phi_z]$, and information about the time of day and weather conditions ($c$).

During the place selection step (Algorithm \ref{alg:place_selection}), the objective is to select unique places and discard duplicate locations from the set of sequences $S$. To achieve this, we determine whether a pose $p$ is new by checking if the linear distance \textbf{dist}$_l$ (Eq. \ref{eq:dist_linear}) or the angular distance \textbf{dist}$_a$ (Eq. \ref{eq:dist_angular}) between $p$ and the poses of all previously selected places exceed two thresholds: $t_{l_{new}}$ for the linear distance and $t_{a_{new}}$ for the angular one. If the condition is met, $p$ is considered the pose of a new place. Algorithm \ref{alg:place_selection} returns a set of poses $P$ representing the unique places in the dataset.

After this, the automatic frame selection can take place. The objective here is to select and associate the RGB images contained in the set of sequences $S$ with the selected places $P$ returned by Algorithm \ref{alg:place_selection}. To achieve this, we determine whether an RGB image $i$ can be associated with a place in $P$ by verifying if the linear distance \textbf{dist}$_l$ (Eq. \ref{eq:dist_linear}) and the angular distance \textbf{dist}$_a$ (Eq. \ref{eq:dist_angular}) between the pose $p$ (corresponding to image $i$) and a place $p'$ in $P$ are below two thresholds: $t_{l_{same}}$ for the linear distance and $t_{a_{same}}$ for the angular one. If the condition is met, $i$ is considered an RGB image that frames the place $p'$. Algorithm \ref{alg:frame_selection} returns a dictionary that associates each place in $P$ with the set of RGB images (and corresponding conditions) that frame it.

\begin{equation} \label{eq:dist_linear}
    \textbf{dist}_l(p, p') = \sqrt{(x-x')^2+(y-y')^2+(z-z')^2}
\end{equation}


\begin{equation} \label{eq:dist_angular}
\begin{split}
    \textbf{dist}_a(p, p') = |(\textbf{angle}(\phi_z) - \textbf{angle}(\phi_z') + \pi) \mathbin{\%} 2\pi - \pi| \\
    \text{where \textbf{angle}}(\phi) =
    \begin{cases} 
    \phi + 2\pi & \text{if } \phi < 0, \\
    \phi & \text{otherwise.}
    \end{cases}
\end{split}
\end{equation}

Note that in our algorithm, the angular distance \textbf{dist}$_a(p, p')$ outlined in Eq. \ref{eq:dist_angular} involves only the rotation around the $z$-axis, which corresponds to the 2D plane, as the rotations around $x$ and $y$ axes are not relevant in this context.

To ensure proper place selection, the thresholds must follow the rules specified in Eq. \ref{rules}. In fact, if the threshold $t_{l_{same}}$ (or $t_{a_{same}}$) is larger than $\frac{t_{l_{new}}}{2}$ (or $\frac{t_{a_{new}}}{2}$), the same RGB image could be associated with two or more places, which is undesirable for place recognition. In other words, the constraints defined in Eq. \ref{rules} ensure that in Algorithm \ref{alg:frame_selection}, no more than one place in $P$ can satisfy the condition of line 9.

\begin{equation} \label{rules}
\begin{split}
    t_{l_{same}} < \frac{t_{l_{new}}}{2}, \text{\hspace{0.2cm}}
    t_{a_{same}} < \frac{t_{a_{new}}}{2}
\end{split}
\end{equation}


\begin{algorithm}[t!]
\caption{Automatic place selection for VPR} \label{alg:place_selection}
\begin{algorithmic}[1]
\Procedure{place\_selection($S$, $ t_{l_{new}}$, $t_{a_{new}}$)}{}
\State $ S \gets $ set of sequences
\State $ t_{l_{new}} \gets $ threshold for linear distance for new place
\State $ t_{a_{new}} \gets $ threshold for angular distance for new place
\State $ P \gets \emptyset$ //set of selected places

\For{\textbf{each} $s$ in $S$}  
    \For{\textbf{each} $(i, p, c)$ in $s$}  
        \If {$\forall p' \in P,$\textbf{dist}$_l(p, p') \geq t_{l_{new}} \vee$ 
        \textbf{dist}$_a(p, p') \geq t_{a_{new}}$} 
        \State $P \gets P \cup \{p\}$
\EndIf

    \EndFor
\EndFor
\State \Return $P$ 
\EndProcedure
\end{algorithmic}
\end{algorithm}

\begin{algorithm}[t!]
\caption{Automatic frames selection for VPR} \label{alg:frame_selection}
\begin{algorithmic}[1]
\Procedure{frames\_selection($S$, $P$, $t_{l_{same}}$, $t_{a_{same}}$)}{}
\State $ S \gets $ set of sequences
\State $ P \gets $ set of selected places
\State $ t_{l_{same}} \gets $ threshold for linear distance for same place
\State $ t_{a_{same}} \gets $ threshold for angular distance for same place
\State $ I \gets \emptyset $ //dictionary \{$p: \{(i, c)\}$\} 
\For{\textbf{each} $s$ in $S$}  
    \For{\textbf{each} $(i, p, c)$ in $s$}  
        \If {$\exists p' \in P,$\textbf{dist}$_l(p, p') < t_{l_{same}} \wedge$ 
        \textbf{dist}$_a(p, p') < t_{a_{same}}$} 
        \State $I[p'] \gets I[p'] \cup \{(i, c)\}$
        \EndIf
    \EndFor
\EndFor
\State \Return $I$ 
\EndProcedure
\end{algorithmic}
\label{alg:places}
\end{algorithm}

To use the acquired dataset for VPR experimentation, we applied the two algorithms to the six paths, thus obtaining a dataset specifically suited for this task. Only the sequences acquired under day/sunny conditions (see Section \ref{preliminary}) were passed to Algorithm \ref{alg:place_selection}, while all the sequences are passed to Algorithm \ref{alg:frame_selection}. Table \ref{tab:parameters} reports the parameters used to create the VPR dataset.

\begin{table}
\centering
\begin{tblr}{
  row{2} = {c},
  cell{1}{1} = {c},
  cell{1}{2} = {c},
  hlines,
}
 $t_{l_{new}}$    &  $t_{a_{new}}$   &  $t_{l_{same}}$   &  $t_{a_{same}}$   \\
100m & 90° & 10m & 20°
\end{tblr}
\caption{Parameters used to create the VPR dataset.}
\label{tab:parameters}
\end{table}

\begin{figure}[ht!]
    \centering

    \subfloat{
        \includegraphics[width=0.48\textwidth]{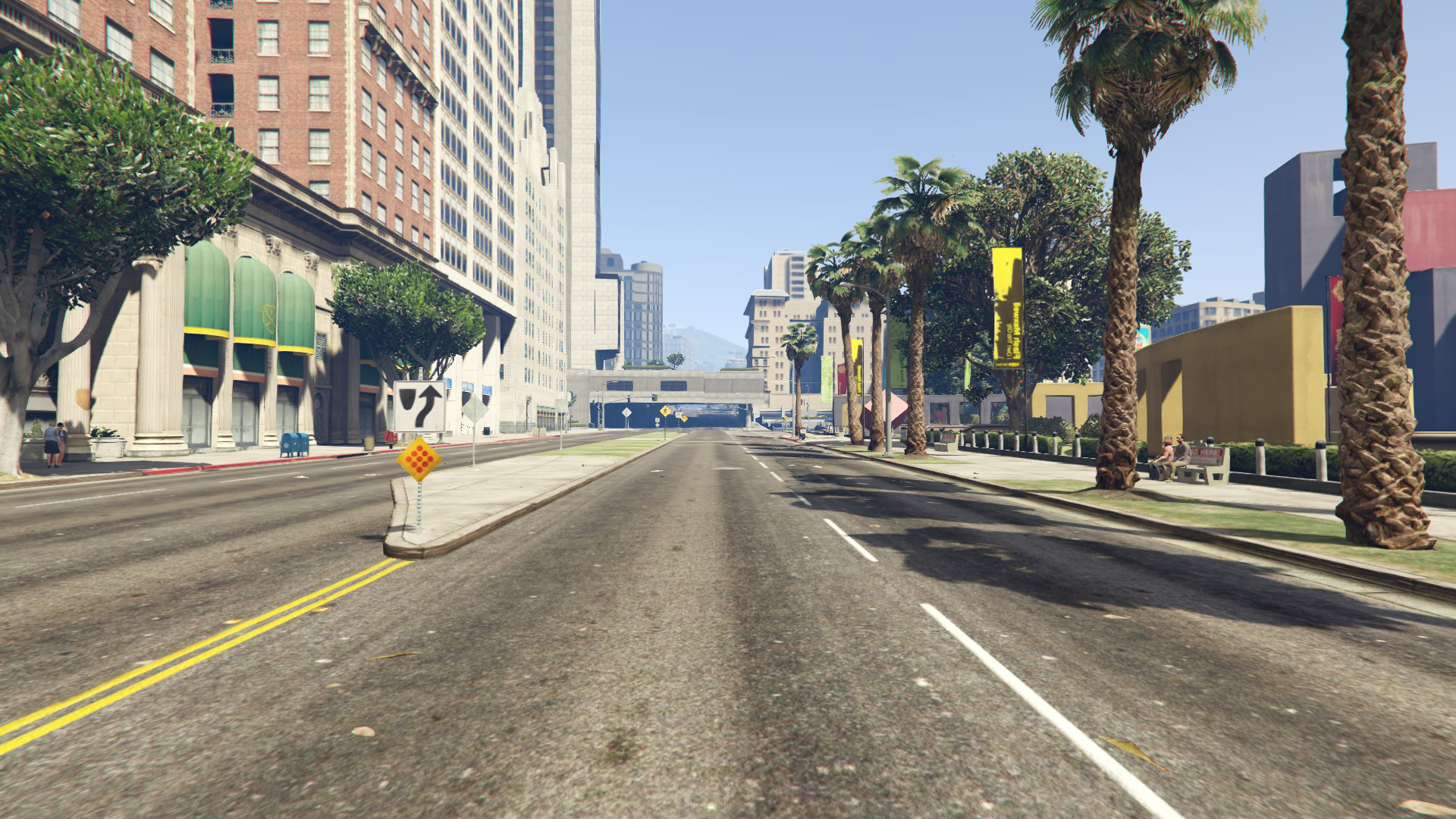}
        \label{fig:vpr_image0}
         \hfill
        \includegraphics[width=0.48\textwidth]{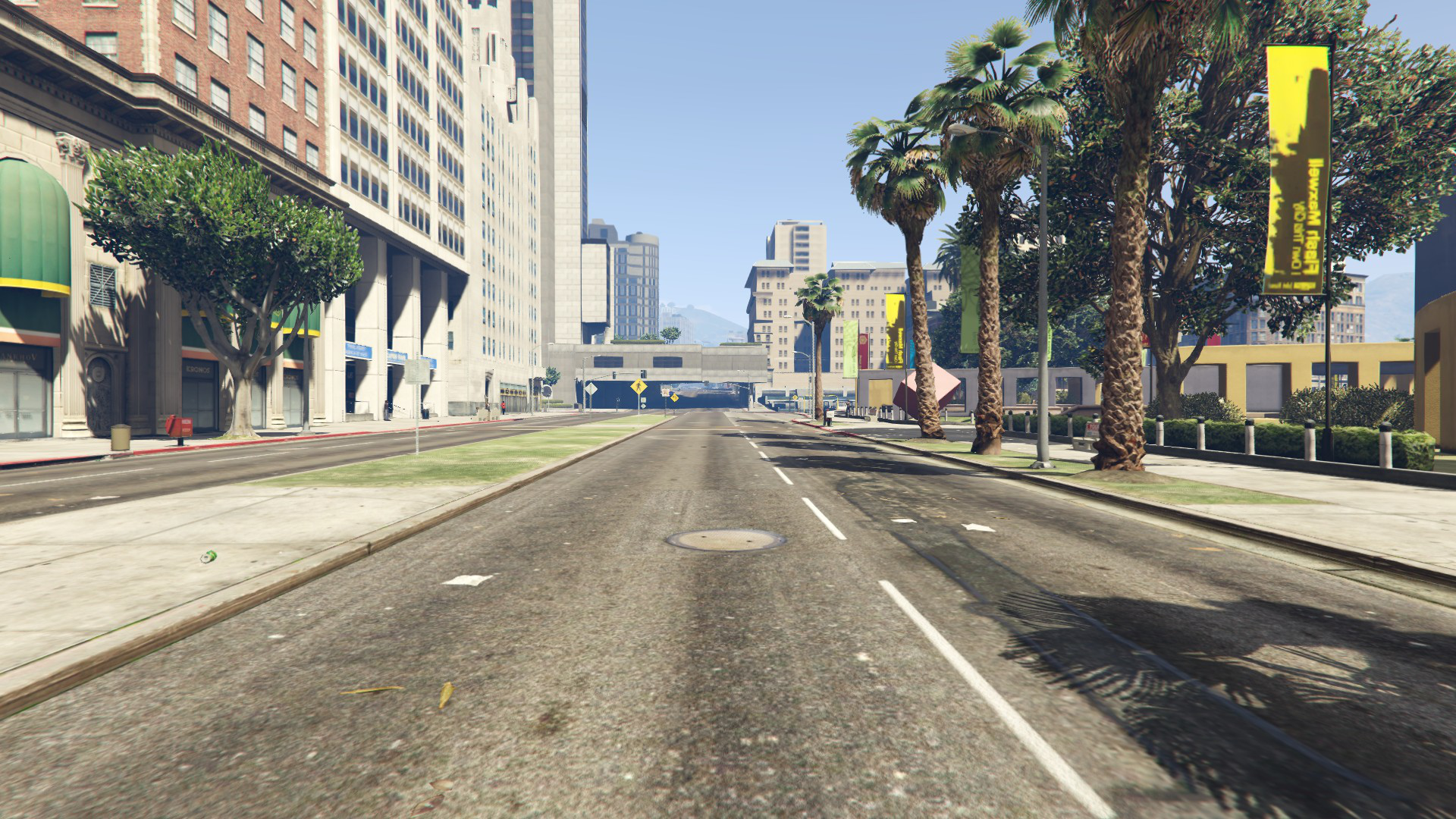}
        \label{fig:vpr_image1}
    }
    
    \vspace{0.3cm}
    
    \subfloat{
        \includegraphics[width=0.48\textwidth]{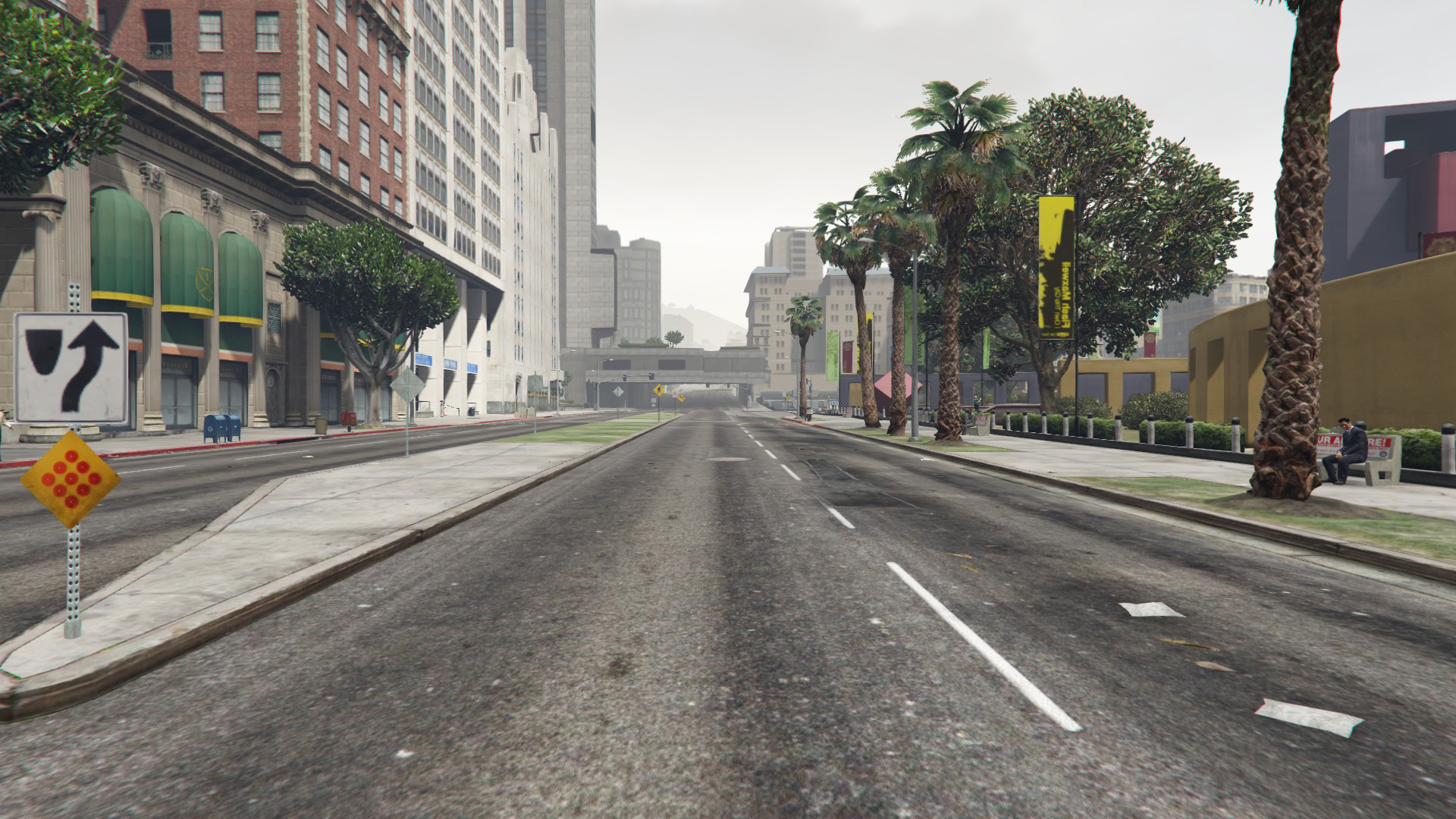}
        \label{fig:vpr_image2}
         \hfill
         \includegraphics[width=0.48\textwidth]{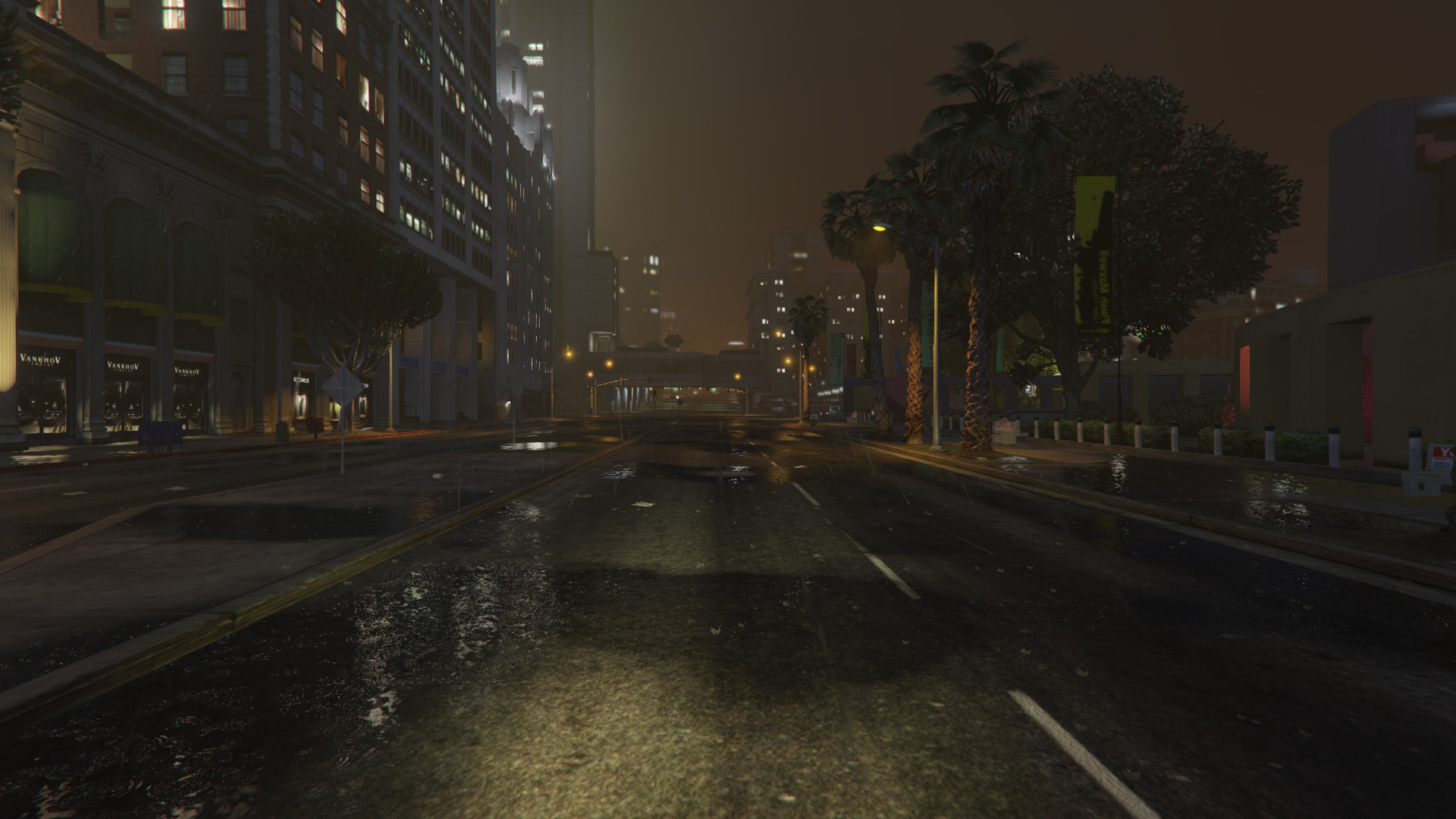}
        \label{fig:vpr_image3}
    }

    \caption{Images associated to the same place by the automatic frame selection algorithm.}
    \label{fig:vpr_image}
\end{figure}

We chose $t_{l_{new}}$ and $t_{a_{new}}$ to achieve a good trade-off between the number of selected locations and their variability. Reducing $t_{l_{new}}$ and $t_{a_{new}}$ results in the algorithm selecting more locations, at the cost of introducing more similar or even identical places. For instance, a building seen from a distance and then re-observed up close may be recognized as the same place by humans, which could affect place recognition. The resulting dataset comprises 20919 RGB images, divided into 326 places, with approximately 65 images for each location, captured in its surroundings under five different conditions. For each location, the associated images consistently frame the same recognizable area, demonstrating that the algorithm performs effectively without human supervision (see Fig. \ref{fig:vpr_image}).

\section{Experiments} \label{experiments}
We conducted several experiments to evaluate the quality of the synthetic data extracted from GTA V, aiming to demonstrate that these data can be used instead of or in addition to real data.
In the following, the results of the experiments on the two tasks selected in this work are reported.

\subsection{Visual Place Recognition}
In the context of VPR, inspired by \cite{tokyo}, we approached the problem as an image retrieval task. Since this study does not aim to propose a new state-of-the-art method for VPR, but rather to demonstrate the usability of GTA V synthetic data for this task, we created a simple but non-trivial scenario: given a day/night image database, when a day (or night) image is presented to the system, it must retrieve the corresponding night (or day) image. Note that, in this scenario, only RGB images are used.

\begin{figure}[ht!]
    \centering
    \subfloat[Alderley]{
        \includegraphics[width=0.48\textwidth]{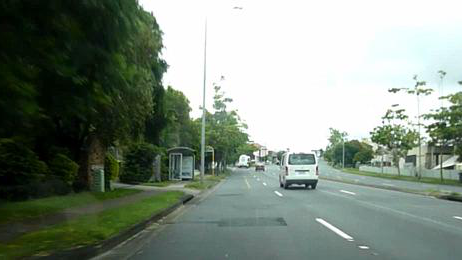}
        \hspace{0.1cm}
        \includegraphics[width=0.48\textwidth]{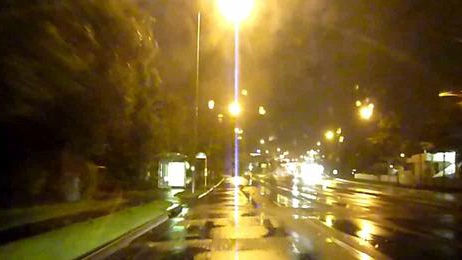}
        \label{fig:alderley}
    }
    \\
    \vspace{0.3cm}
    \subfloat[Tokyo 24/7]{
        \includegraphics[width=0.48\textwidth]{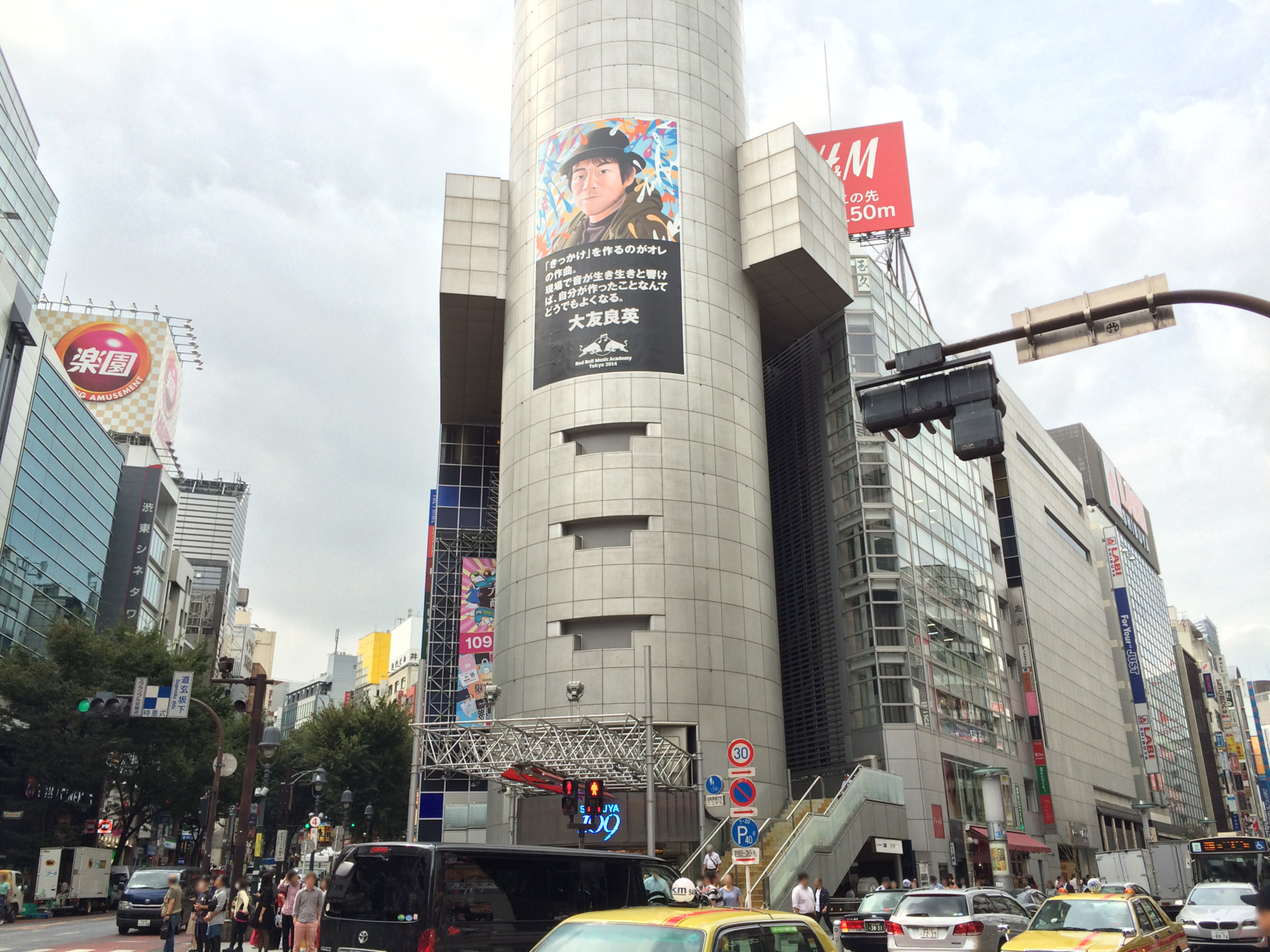}
        \hspace{0.1cm}
        \includegraphics[width=0.48\textwidth]{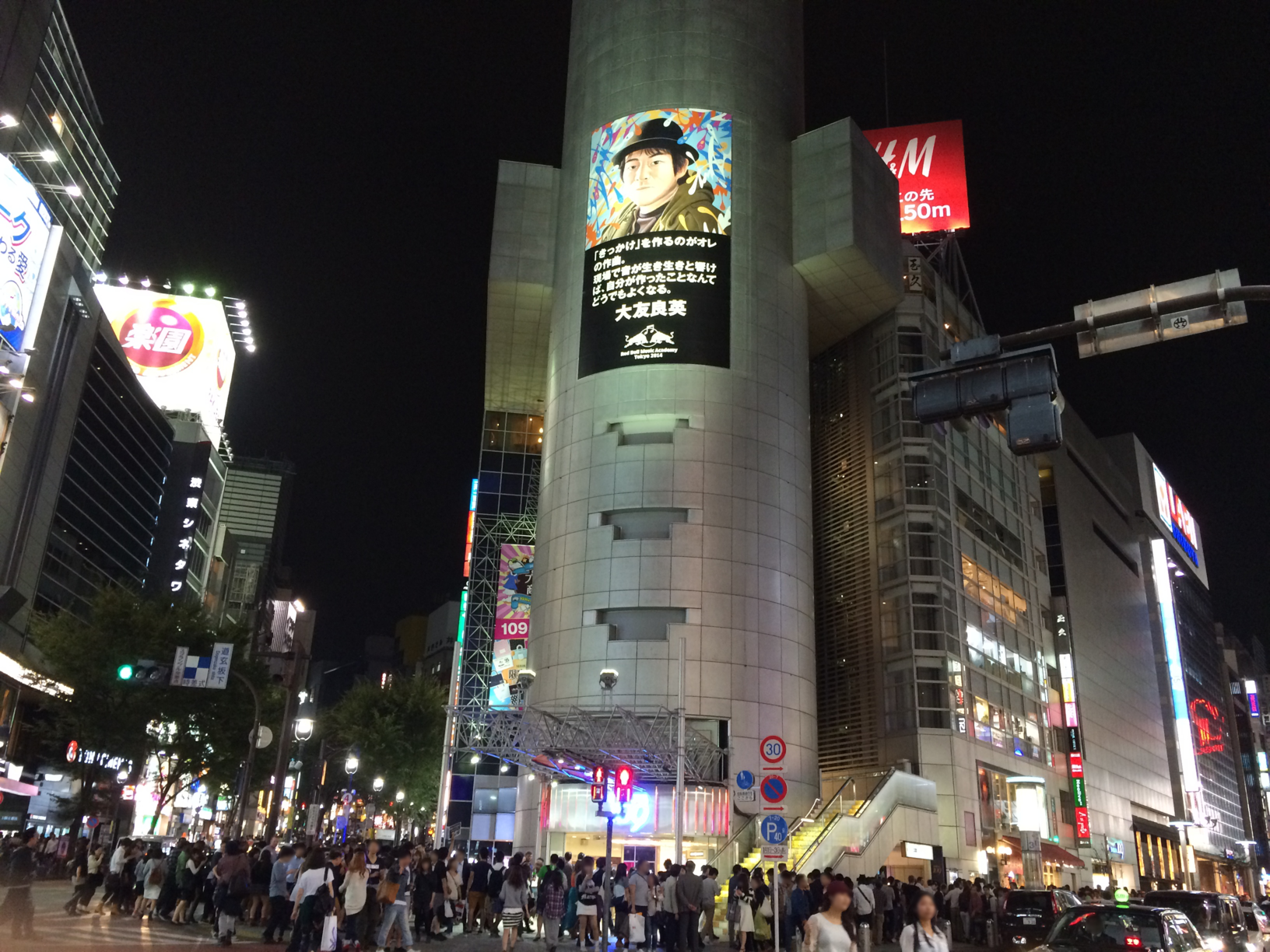}
        \label{fig:tokyo}
    }
    \\
    \vspace{0.3cm}
    \subfloat[GTA V]{
        \includegraphics[width=0.48\textwidth]{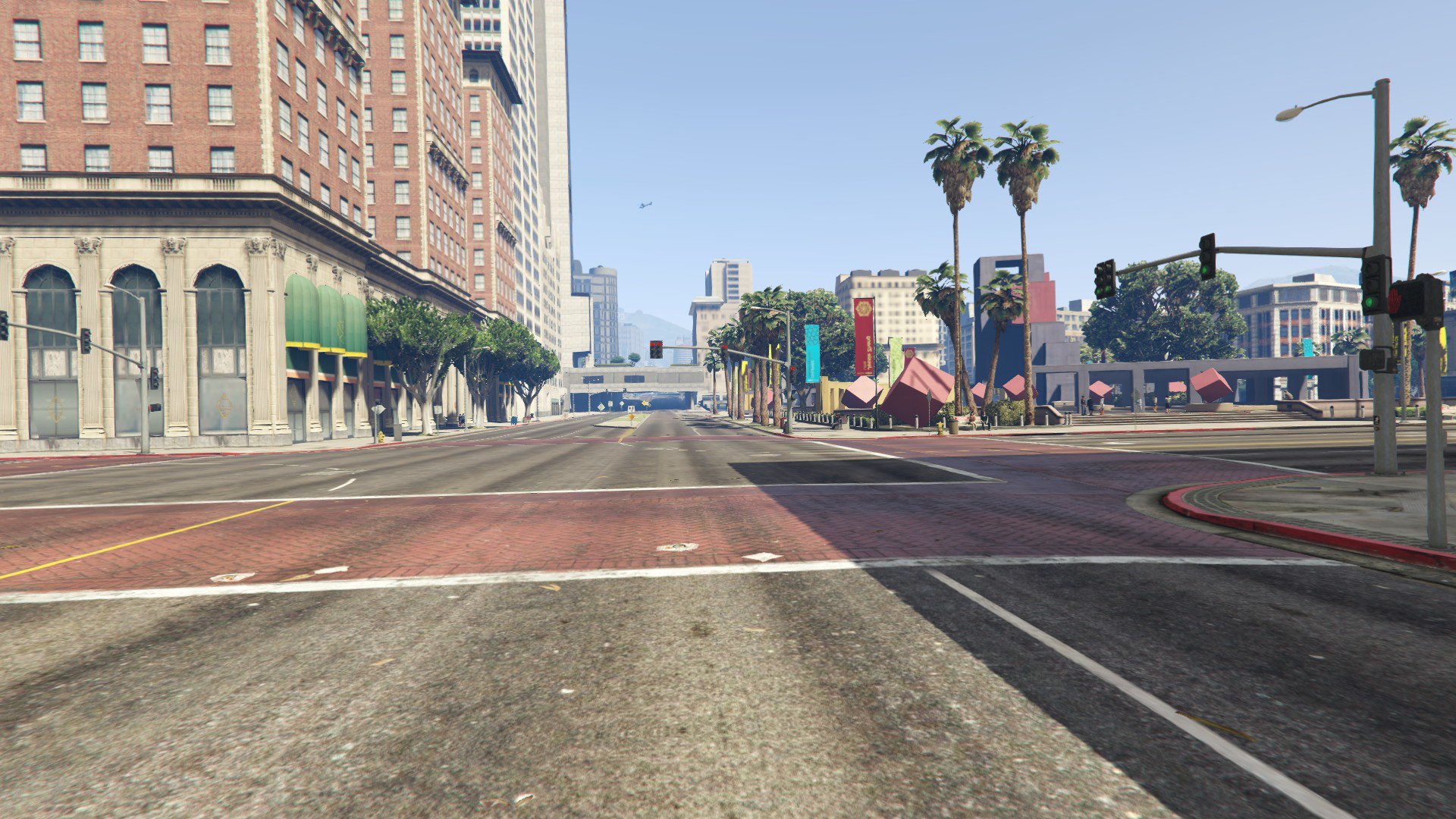}
        \hspace{0.1cm}
        \includegraphics[width=0.48\textwidth]{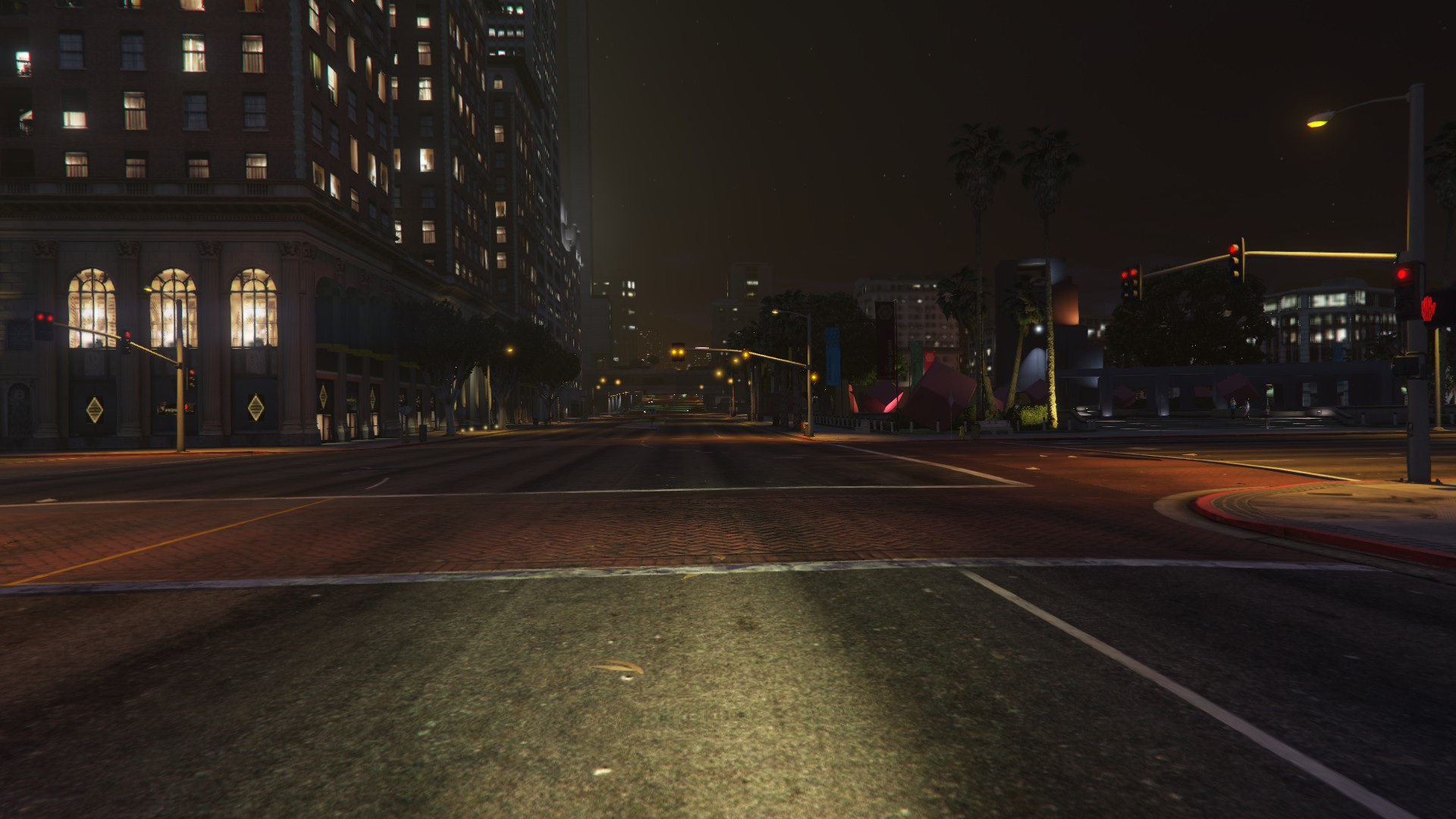}
        \label{fig:gta}
    }
    \caption{Example of images from the three datasets, captured under different light/weather conditions.}
    \label{fig:datasets_images}
\end{figure}

For these experiments, we used three datasets:
\begin{itemize}
    \item Alderley \cite{alderley}, a real dataset of paired day/night RGB images captured by a camera on top of a car in an urban environment. For the VPR task, 125 pairs of frames have been manually selected, to obtain enough diversity between places (see Fig. \ref{fig:alderley}).
     \item Tokyo 24/7 query V3 \cite{tokyo}, a subset of the Tokyo 24/7 dataset reported in Section \ref{related}, a real dataset of 375 places with day/night images captured in Tokyo. This dataset is specifically designed for VPR, therefore no frame selection is required (see Fig. \ref{fig:tokyo}).
    \item GTA V, the VPR synthetic dataset acquired in this work using GTA V (see Section \ref{capture}). For each place, the automatic algorithm described \ref{vpr_dataset} is used to select frames for VPR and to pair day/night images of a place. A total of 326 different places have been automatically selected.
   
\end{itemize}
Fig. \ref{fig:datasets_images} shows image pairs from the three datasets.

To ensure a fair comparison of the results obtained using the Alderley or GTA V datasets during training, only the first 125 locations automatically selected by GTA are used, providing an equal number of places from both real and synthetic datasets. The Tokyo 24/7 dataset was not used for training the system, but only for testing.

\begin{figure}[ht!]
    \centering
    \includegraphics[width=\textwidth]{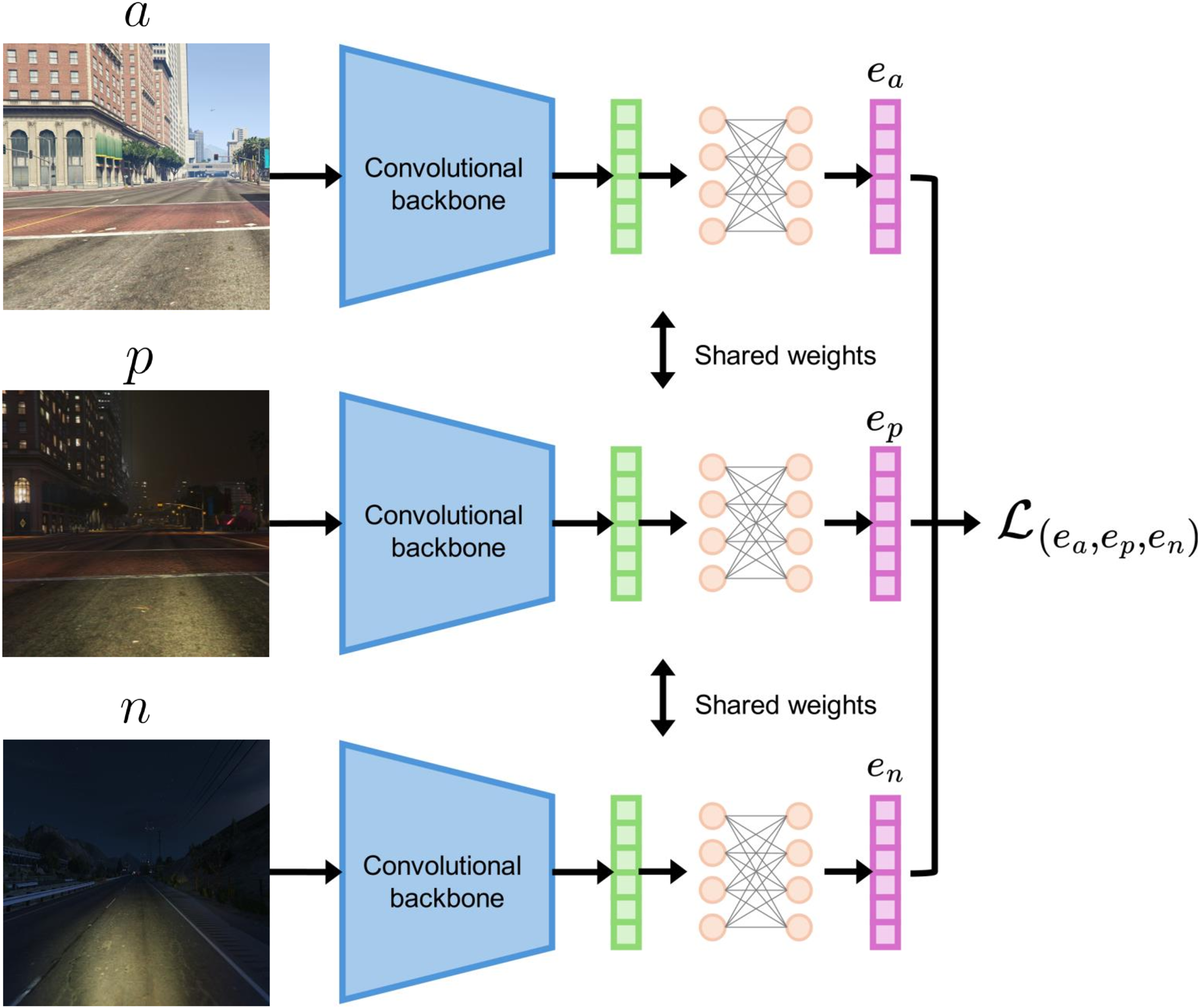}
    \caption{Training scheme for the VPR experiments.}
    \label{fig:training}
\end{figure}

We trained a deep neural network (see Fig. \ref{fig:training}) to perform the task. Specifically, we used a triplet neural network consisting of three replicas of the same network with shared weights. The network is composed of a convolutional backbone (ResNet50\cite{resnet50} in our case) pre-trained on ImageNet \cite{imagenet} with frozen weights serving as the feature extractor, and a final Multi-Layer Perceptron (MLP) comprising two fully connected layers to project the features onto a new latent space. The purpose of the final MLP is to bring the latent representations of the positive pairs (day image with the corresponding night and vice versa) closer together in latent space, while moving the latent representations of the negative pairs farther apart. The metric used to compute the distance between the embeddings is the classic cosine distance; during training, the cosine distance between positive pairs must be minimized, while the cosine distance between negative pairs must be maximized. 

To train the neural network in this manner, triplets consisting of an anchor image $a$, a positive image $p$, and a randomly chosen negative image $n$ are created. The neural network is used to compute the embeddings of the three images, obtaining $e_a, e_p$ and $e_n$ respectively.

Given the distance function $d$ (cosine distance) and a margin $\epsilon$, the loss function is computed as shown in Eq. \ref{eq:triplet_loss}.

\begin{equation} \label{eq:triplet_loss}
    \mathcal{L}_{(e_a, e_p, e_n)} = max(d(e_a, e_p) - d(e_a, e_n) + \epsilon, 0)
\end{equation}

which corresponds to

\begin{equation} \label{eq:triplet_loss_mean}
\mathcal{L} = \frac{1}{k}\sum^k_{i=1} \mathcal{L}_{(e_a^i, e_p^i, e_n^i)}
\end{equation}

on each minibatch with $k$ samples.

In this context, if $e_a^i$ is a day image, both $e_p^i$ and $e_n^i$ must be night images and vice versa. 

Alderley and GTA V images (which have an aspect ratio of 16:9) have been randomly cropped during training using a window of aspect ratio 4:3 to ensure the same aspect ratio of the images from Tokyo 24/7. The random crop is not only useful to avoid different stretches between different datasets but also to introduce further variability in the data during training, since the position of the window is computed on the fly for each sample. Subsequently, all images were resized to 224x224 pixels to be compatible with the network.

We performed three experiments for VPR: (i) using the Alderley as the training set (only real images), (ii) using the GTA V as the training set (only synthetic images) and (iii) using both the Alderley and GTA V as the training set (real and synthetic images). In the third experiment, negative images ($n_i$) were selected from the same dataset as the positive images ($a_i$), ensuring that if $a_i$ was real, then $n_i$ was also real, and vice versa. The Tokyo 24/7 dataset was always used as the test set.

\begin{figure}[ht!]
    \centering
    \begin{subfigure}[b]{0.48\textwidth}
        \centering
        \includegraphics[width=\textwidth]{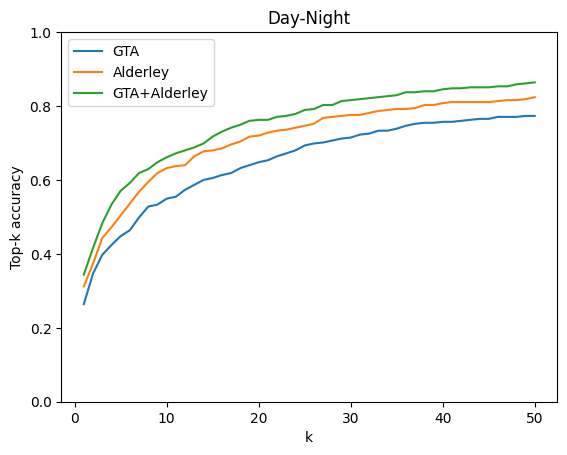}
        \caption{}
        \label{fig:total_day_night}
    \end{subfigure}
    \hfill
    \begin{subfigure}[b]{0.48\textwidth}
        \centering
        \includegraphics[width=\textwidth]{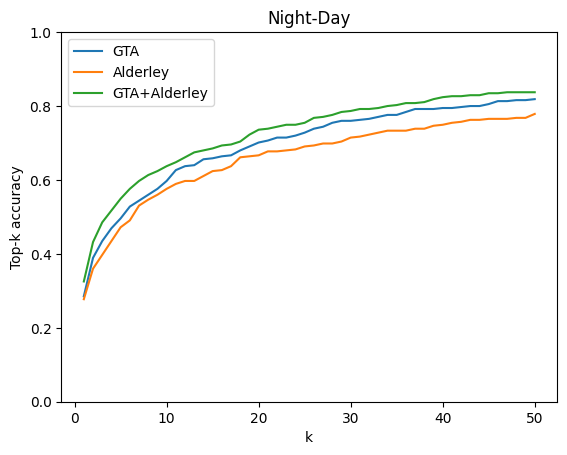}
        \caption{}
        \label{fig:total_night_day}
    \end{subfigure}
    \caption{Results of the VPR experiments. On the left (a) shows the recall at different $k$ using day images of Tokyo 24/7 as query and night images as database, on the right (b) reports the results of the opposite setting, using night images as query and day images as database.}
    \label{fig:vpr_results}
\end{figure}

To validate the experiments, we approached the problem as an image retrieval task. We used Tokyo 24/7 day (night) images as database and each night (day) image of Tokyo 24/7 is presented to the network as query, to retrieve the most similar day (night) images. 
The evaluation metric employed was top-$k$ accuracy, which measures whether the correct prediction $y_i$ is among the top-$k$ predictions, and we varied the values of $k$ to assess how the recall changes when $k$ increases (see Fig. \ref{fig:vpr_results}).

The results obtained using GTA V and Alderley as training sets are comparable, with GTA V even outperforming Alderley in some instances, as illustrated in Fig. \ref{fig:total_night_day}. This demonstrates that synthetic RGB images from GTA V are sufficient to train a neural network for this task. Moreover, the results show that combining synthetic and real data during training further improves retrieval performance. Although this may seem intuitive, it has significant implications for the dataset acquisition process. A new dataset can potentially be created using a small number of real images, whose acquisition requires time, energy, and possibly cost, while the majority of the dataset can be composed of synthetic images, which enrich the real data and increase the number of samples at approximately no cost, but resulting in improved system performance.

\subsection{SLAM}
\begin{figure}[ht!]
    \centering
    \begin{subfigure}[b]{0.32\textwidth}
        \centering
        \includegraphics[width=\textwidth]{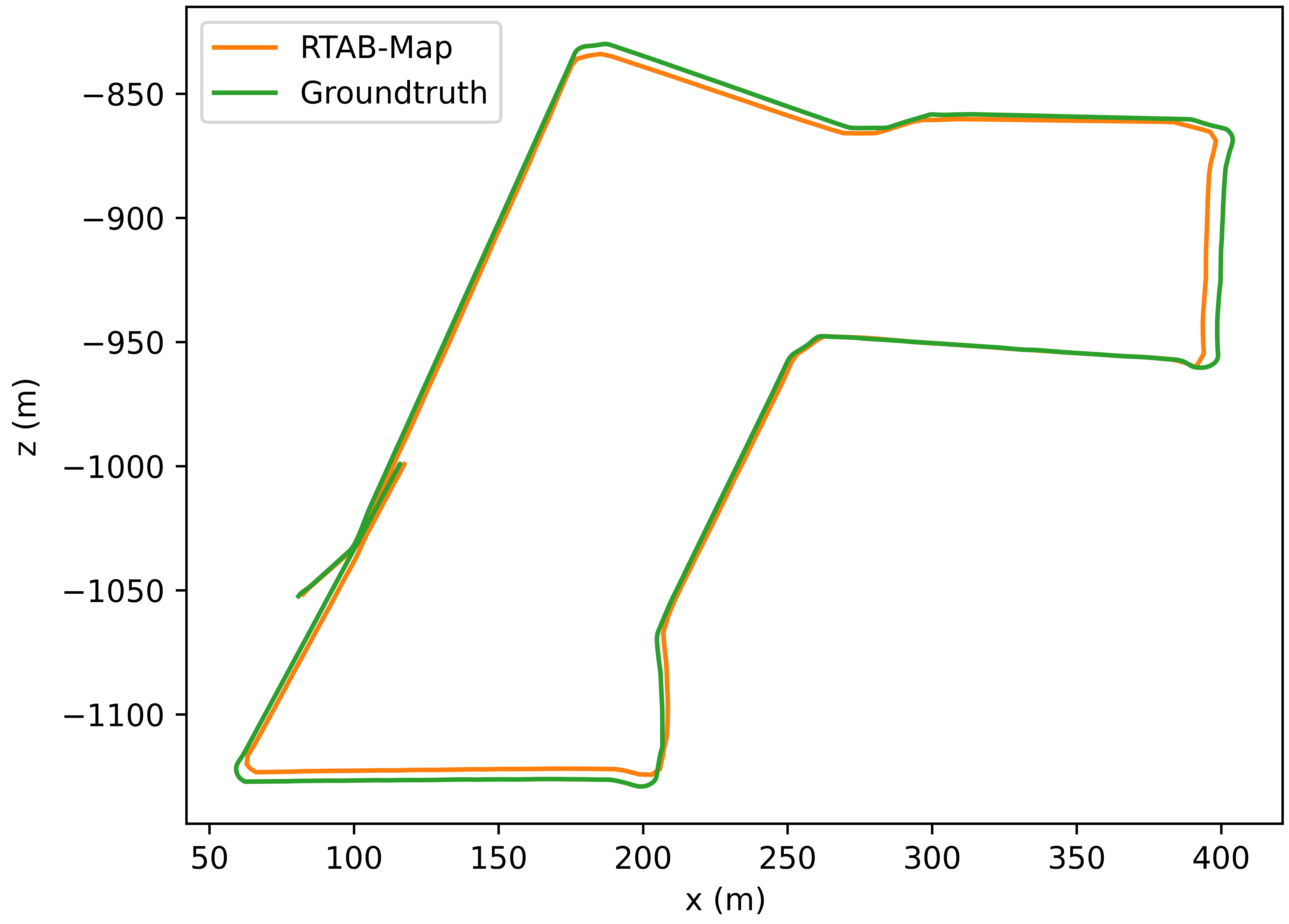}
        \caption{}
        \label{fig:legionsquare_slam}
    \end{subfigure}
    \hfill
    \begin{subfigure}[b]{0.32\textwidth}
        \centering
        \includegraphics[width=\textwidth]{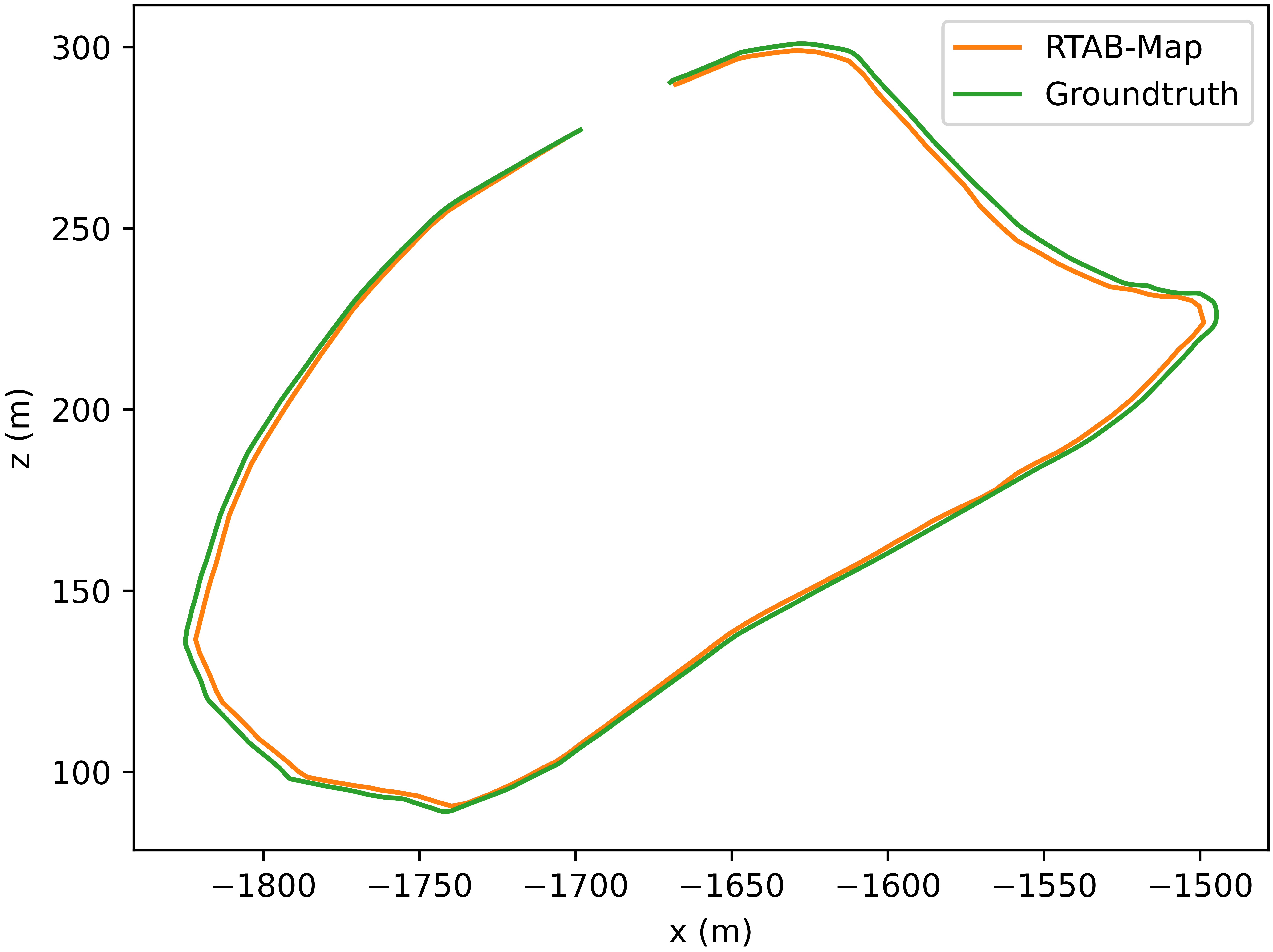}
        \caption{}
        \label{fig:stadium_slam}
    \end{subfigure}
    \hfill
    \begin{subfigure}[b]{0.32\textwidth}
        \centering
        \includegraphics[width=\textwidth]{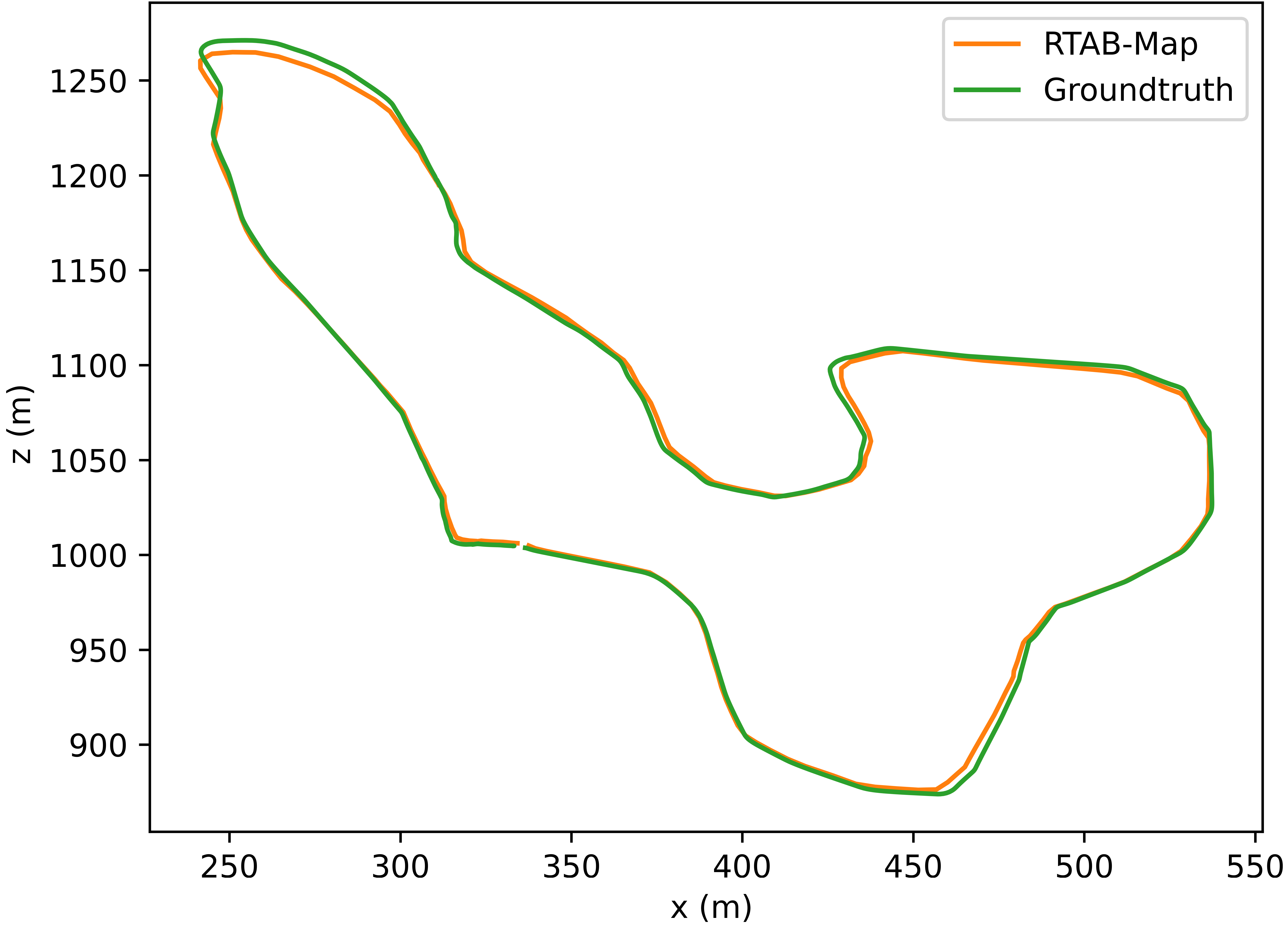}
        \caption{}
        \label{fig:vinewooda_slam}
    \end{subfigure}
    
    \vskip\baselineskip
    
    \begin{subfigure}[b]{0.32\textwidth}
        \centering
        \includegraphics[width=\textwidth]{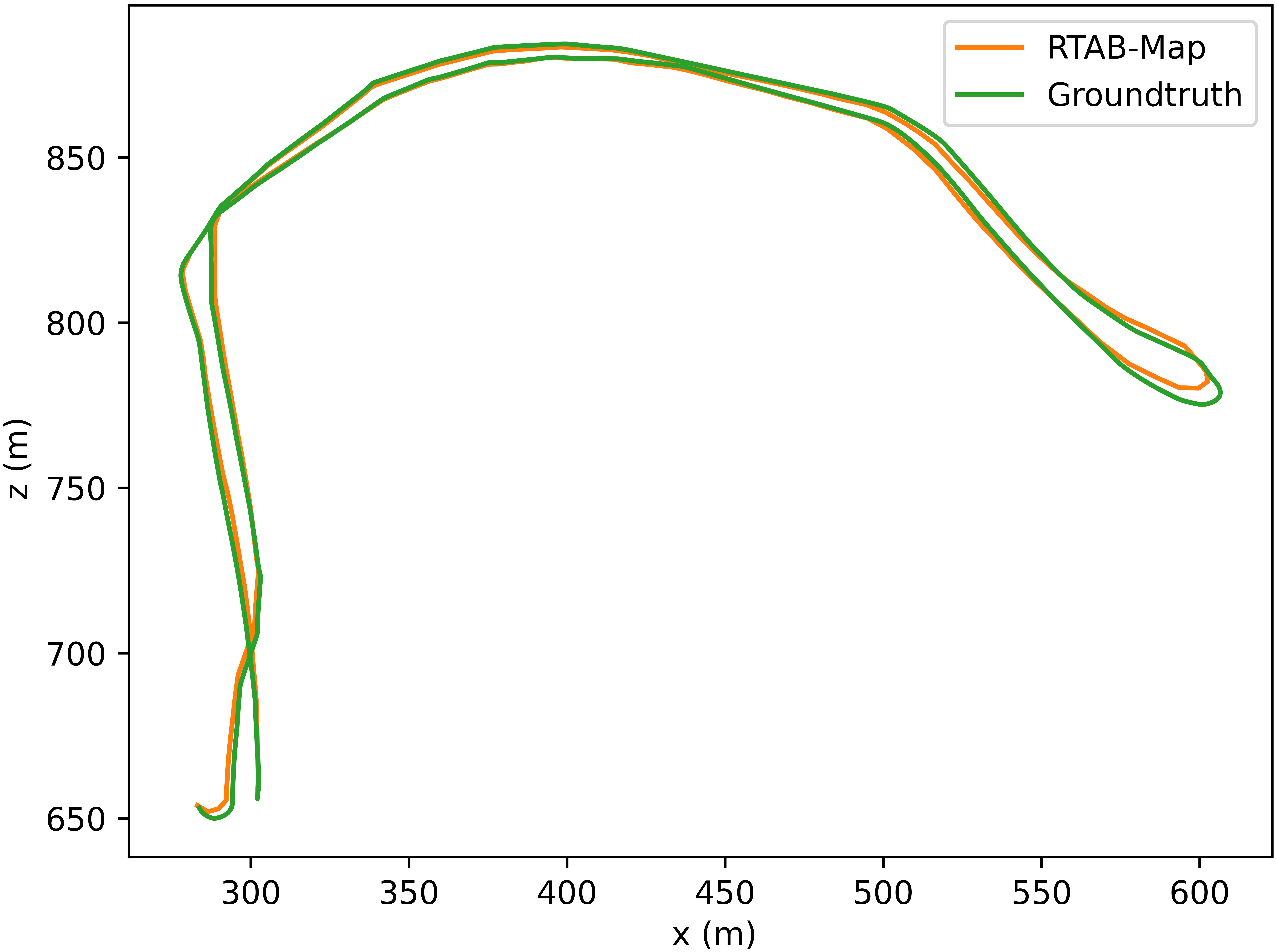}
        \vspace{0.2cm}
        \caption{}
        \label{fig:vinewoodb_slam}
    \end{subfigure}    
    \begin{subfigure}[b]{0.32\textwidth}
        \centering
        \includegraphics[width=\textwidth]{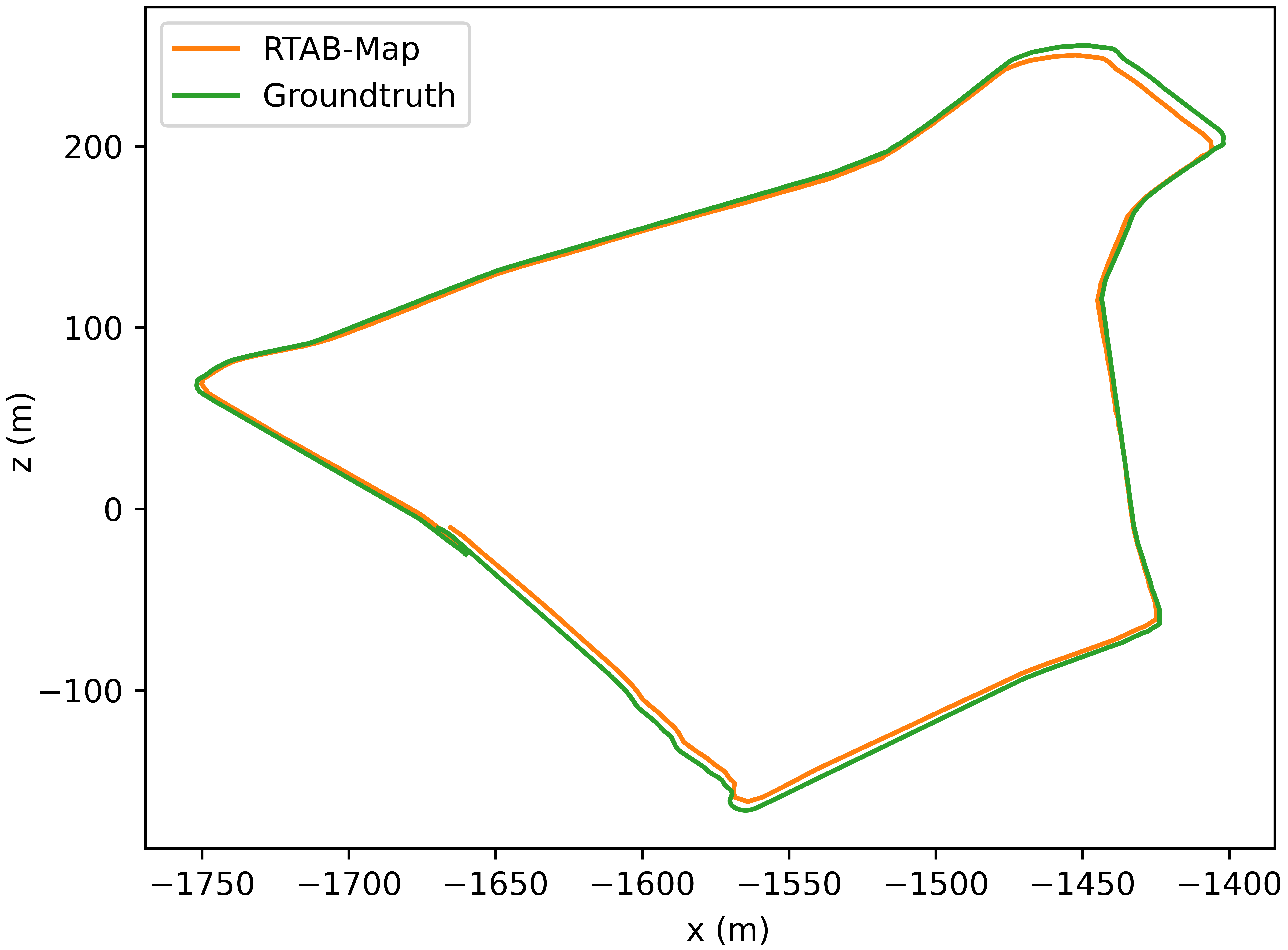}
        \vspace{0.2cm}
        \caption{}
        \label{fig:westeclipse_slam}
    \end{subfigure}
    \begin{subfigure}[b]{0.32\textwidth}
        \centering
        \includegraphics[width=\textwidth]{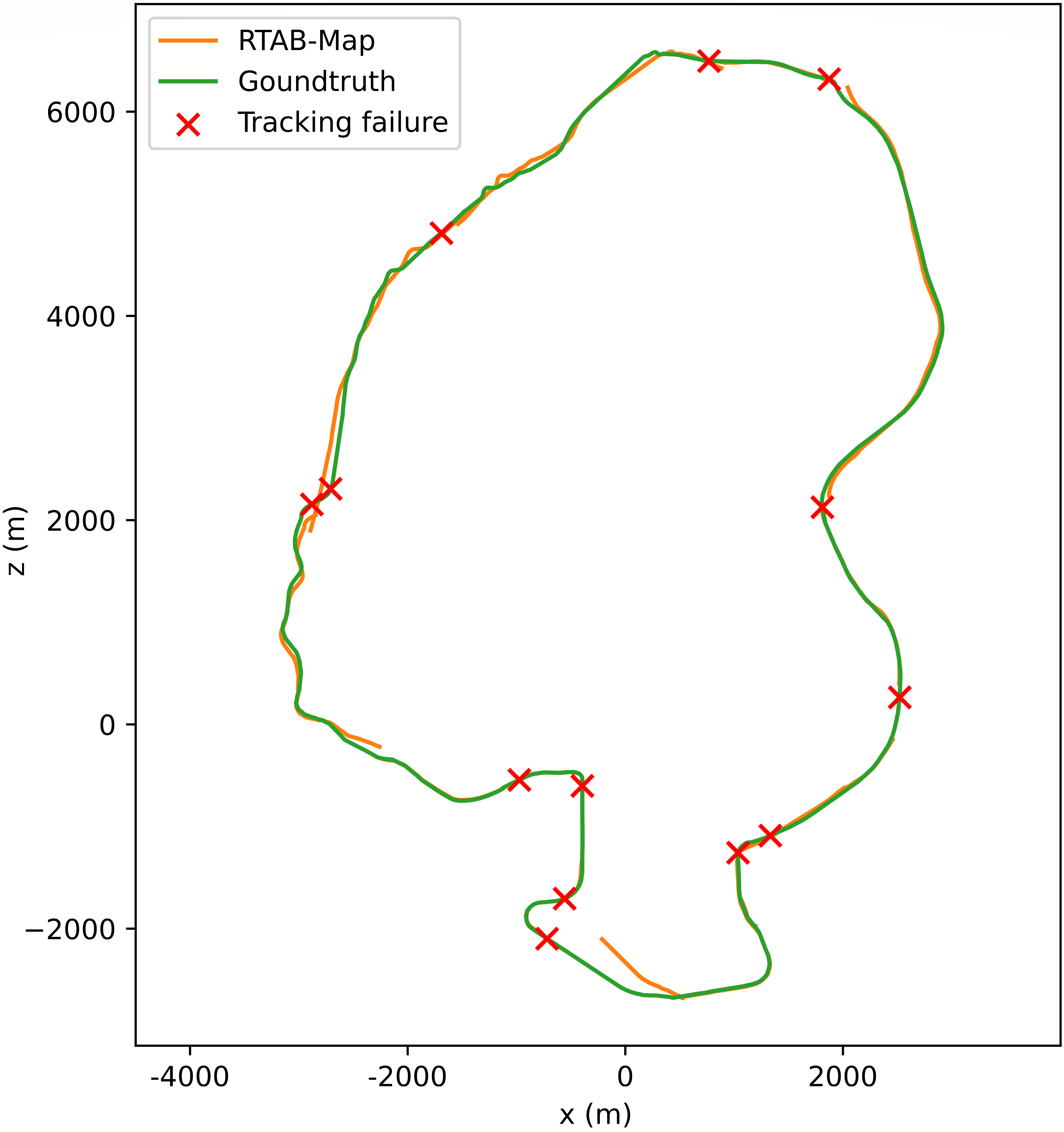}
        \caption{}
        \label{fig:longring_slam}
    \end{subfigure}
    
    \caption{Paths estimated by RTAB-Map compared to the ground truth. In order, (a) is LegionSquare, (b) is Stadium, (c) is VinewoodA, (d) is VinewoodB, (e) is WestEclipseBoulevard and (f) is the LongRing. The red crosses on the LongRing path highlight where the tracking is lost.}
    \label{fig:slam_results}
\end{figure}

An additional experiment was conducted in the context of visual SLAM, a more challenging task that requires precise sensor acquisition. We evaluated the paths detailed in Section \ref{preliminary} using RTAB-Map \cite{rtabmap}, a well-known graph-based SLAM approach.
To compute the itinerary traced by the car, RTAB-Map requires RGB-D images, so depth images are also used in this experiment. This is crucial because, to achieve accurate odometry and correctly detect loops, the depth images must be very precise, otherwise the results obtained from the ground truth of the poses would be completely wrong.
Fig. \ref{fig:slam_results} shows the trajectories estimated by RTAB-Map compared to the ground truths. The estimated paths are extremely accurate, with only minor physiological inaccuracies of SLAM. This highlights that GTA V data can be used for this task, proving that RGB and depth images are as accurate as, if not more accurate than, real-world data. While this represents a significant strength, it also poses a disadvantage. In real-world scenarios, robots equipped with real sensors have to deal with noise and measurement errors. Consequently, especially for deep learning-based solutions, the optimal approach might be to combine real-world data with synthetic data from GTA V, or introduce noise in the synthetic depth images, to obtain a dataset that incorporates the variability and imperfections inherent in real-world data.

The metrics used to measure the goodness of the results are based on the accuracy of the localization, which typically also reflects the accuracy of the map. The most common metric is the Absolute Trajectory Error (ATE) \cite{ate} which measures the absolute distances between the estimated and the ground truth trajectory, but also other metrics exist \cite{ate, openloris}. Since both trajectories can be defined in arbitrary coordinate frames, alignment is necessary before comparison. This alignment is typically achieved using Horn's method \cite{horn}, which computes the rigid-body transformation $S$ that provides the least-squares solution for mapping the estimated trajectory $P_{1:n}$ onto the ground truth trajectory $Q_{1:n}$. The absolute trajectory error at time step $i$ is then calculated as:

\begin{equation}
    F_i = Q^{-1}_iSP_i
\end{equation}

ATE evaluates the root mean squared error (RMSE) over all time indices of the translation components as follows:

\begin{equation}
    RMSE(F_{1:n}) = \sqrt{\frac{1}{n} \sum_{i=1}^{n} \| trans(F_i) \|^2 }
\end{equation}

Table \ref{tab:ate} reports the ATE for the five sequences. The errors observed are relatively small, perfectly in line with the expected errors for paths of comparable size with real data.

\begin{table}[ht!] 
\centering
\begin{tabular}{cc} 
\hline
\textbf{Sequence}    & \textbf{ATE (m)} \\ \hline
LegionSquare         & 1.49             \\ 
Stadium              & 1.38             \\ 
VinewoodA            & 1.16             \\ 
VinewoodB            & 0.94             \\ 
WestEclipseBoulevard & 1.47             \\ \hline
\end{tabular}
\caption{Absolute Trajectory Errors for the five sequences run on RTAB-Map.}
\label{tab:ate}
\end{table}

In this experiment, RTAB-Map was unable to compute the entire itinerary of the LongRing path due to poor odometry and tracking failures in several segments of the sequence. This further emphasizes the need for such a dataset: exploring a wide range of locations, including those with minimal texture or where lighting conditions change (e.g. a car tunnel), requires robust tracking and odometry techniques. Common SLAM solutions, which rely solely on images, are insufficient to handle such scenarios. Therefore, a sequence like LongRing highlights the need for more robust SLAM systems to deal with real-world, long-term acquisitions under varying conditions and in untextured environments.
In Fig. \ref{fig:slam_results}, the LongRing path was created by aligning with the ICP algorithm the subparts obtained from RTAB-Map. The aligned paths are as accurate as the other sequences, further highlighting data from GTA V is suitable for this task. The red crosses in the image highlight where the tracking is lost; in these points, the images (i) are mainly composed of untextured regions (e.g., the sky, a uniform road), causing common algorithms to fail in detecting and tracking a sufficient number of features to estimate accurate odometry, or (ii) exhibit abrupt changes in illumination conditions, preventing feature tracking (e.g. when entering or exiting a car tunnel).

\section{Conclusions and Future Work} \label{conclusion}
In this study, we introduced a novel synthetic dataset acquired from the famous video game GTA V. We evaluated the dataset on two tasks where the use of synthetic data is uncommon: VPR and SLAM. The experiments demonstrated that data from GTA V can be used as a substitute for, or in conjunction with, real data to improve performance in both tasks. Furthermore, the SLAM experiments revealed the limitations of common visual SLAM systems, which are not able to deal with exploration in the dark or in untextured environments. In the context of VPR, we also proposed a simple but effective algorithm for generating a VPR dataset from image sequences and poses ground truth without requiring human supervision.
The proposed pipeline facilitates the creation of large datasets under controllable conditions that are challenging to capture in the real world and at almost no cost. This methodology allows the community to rapidly enrich the plethora of existing datasets for each task with new data acquired under novel conditions or scenarios that are difficult to acquire in real-world settings. 
Future work will focus on the creation of novel datasets of unprecedented size for each task reported in \ref{tasks}, using GTA V and the tools and pipeline described in \ref{capture}. The datasets will be generated under new conditions and scenarios that are not typically considered by the community.
\printbibliography
\end{document}